\documentclass[sigconf,preprint]{acmart}
\AtBeginDocument{%
  }
\setcopyright{acmlicensed}
\copyrightyear{2018}
\acmYear{2018}
\acmDOI{XXXXXXX.XXXXXXX}
\acmConference[Preprint]{Make sure to enter the correct
  conference title from your rights confirmation email}{June, 2026}{Sydney, Australia}
\acmISBN{978-1-4503-XXXX-X/2018/06}

\usepackage{hyperref}
\usepackage{url}
\usepackage{booktabs}       
\usepackage{amsfonts}       
\usepackage{nicefrac}       
\usepackage{microtype}      
\usepackage{xcolor}         
\usepackage{graphicx}
\usepackage{multirow}
\usepackage{makecell}
\usepackage{amsmath}
\usepackage{subcaption}
\usepackage{adjustbox}
\usepackage{wrapfig}
\usepackage{mathtools} 
\usepackage{arydshln}
\usepackage{dblfloatfix}

\begin{document}

\title{HiT-JEPA: A Hierarchical Self-supervised Trajectory Embedding Framework for Similarity Computation}


\author{Lihuan Li}
\affiliation{%
  \institution{University of New South Wales}
  \city{Sydney}
  \country{Australia}}
\email{lihuan.li@unsw.edu.au}

\author{Hao Xue}
\affiliation{%
  \institution{Hong Kong University of Science and Technology (Guangzhou)}
  \city{Guangzhou}
  \country{China}
}
\email{haoxue@hkust-gz.edu.cn}

\author{Shuang Ao}
\affiliation{%
 \institution{University of New South Wales}
 \city{Sydney}
 \country{Australia}}
\email{shuang.ao@unsw.edu.au}

\author{Yang Song}
\affiliation{%
 \institution{University of New South Wales}
 \city{Sydney}
 \country{Australia}}
\email{yang.song1@unsw.edu.au}

\author{Flora Salim}
\affiliation{%
 \institution{University of New South Wales}
 \city{Sydney}
 \country{Australia}}
\email{flora.salim@unsw.edu.au}


\begin{abstract}
The representation of urban trajectory data plays a critical role in effectively analyzing spatial movement patterns. Despite considerable progress, the challenge of designing trajectory representations that can capture diverse and complementary information remains an open research problem. Existing methods struggle to incorporate trajectory fine-grained details and high-level summary in a single model, limiting their ability to attend to both long-term dependencies while preserving local nuances. To address this, we propose HiT-JEPA (\textbf{H}ierarchical \textbf{I}nteractions of \textbf{T}rajectory Semantics via a \textbf{J}oint \textbf{E}mbedding \textbf{P}redictive \textbf{A}rchitecture), a unified framework for learning multi-scale urban trajectory representations across semantic abstraction levels. HiT-JEPA adopts a three-layer hierarchy that progressively captures point-level fine-grained details, intermediate patterns, and high-level trajectory abstractions, enabling the model to integrate both local dynamics and global semantics in one coherent structure. Extensive experiments on multiple real-world datasets for trajectory similarity computation show that HiT-JEPA's hierarchical design yields richer, multi-scale representations. Code is available at: \url{https://anonymous.4open.science/r/HiT-JEPA}.
\end{abstract}

\ccsdesc[500]{Networks~Location based services}
\ccsdesc[300]{Applied computing~Transportation}
\ccsdesc[500]{Information systems~Spatial-temporal systems}

\keywords{trajectory similarity computation, hierarchical self-supervised learning, transformers, joint embedding predictive architecture}

\received{20 February 2007}
\received[revised]{12 March 2009}
\received[accepted]{5 June 2009}

\maketitle

\section{Introduction} \label{introduction}
With the widespread use of location-aware devices, trajectory data is now produced at an unprecedented rate~\cite{zhu2024unitraj,qian2024context}. Effectively representing trajectory data powers critical applications ranging from urban computing applications, such as travel time estimation~\cite{chen2022interpreting,chen2021robust,lin2023pre}, trajectory clustering~\cite{fang20212,yao2024deep,bai2020adaptive}, and traffic analysis~\cite{yu2017spatio}. Trajectories exhibit multi-scale attributes, ranging from short-term local transitions (e.g., turns and stops) to long-term strategic pathways or routines, whereas capturing both the fine-grained point-level details of individual trajectories and higher-level semantic patterns of mobility behavior within a unified framework is challenging. This necessitates a representation learning model that accommodates this complexity.

\begin{figure}[!ht]
\centering
\includegraphics[width=\linewidth]{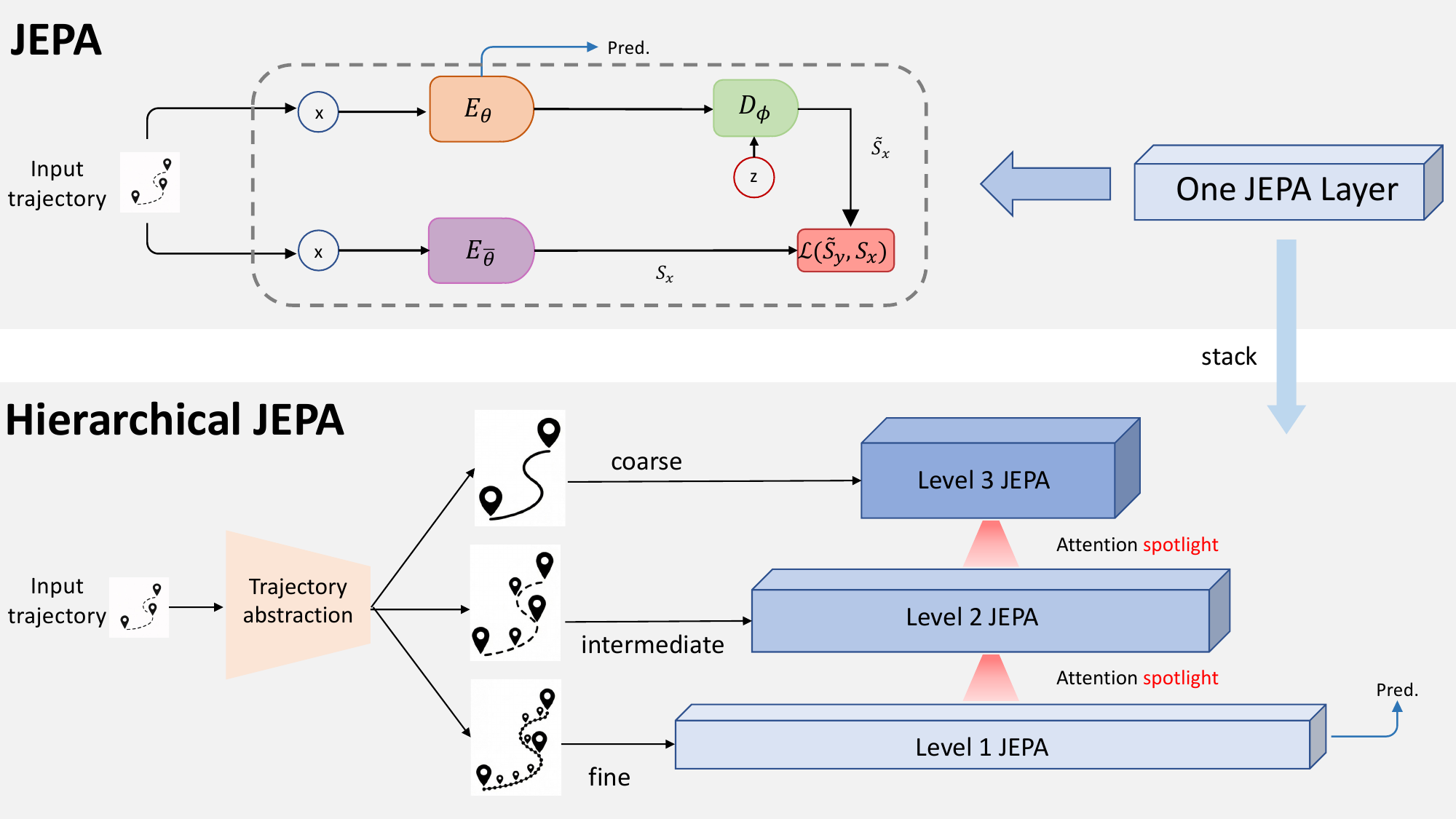}
\caption{Structural comparisons between JEPA and Hierarchical JEPA.}
\label{fig:intro}
\end{figure}

Early trajectory analysis methods (heuristic methods)~\cite{alt1995computing,chen2004marriage,chen2005robust,yi1998efficient} relied on handcrafted similarity measures and point-matching heuristics. Recently, deep-learning-based approaches have been applied to learn low-dimensional trajectory embeddings, alleviating the need for manual feature engineering~\cite{yang2024simformer,yao2019computing,yang2021t3s}. Self-supervised learning frameworks~\cite{li2018deep,cao2021accurate}, especially contrastive learning, further advanced trajectory representation learning by leveraging large unlabeled datasets~\cite{chang2023contrastive,liu2022cstrm,li2024clear}. However, these deep learning models usually generate a single scale embedding of an entire trajectory and cannot integrate different semantic levels, i.e., they often neglect fine-grained point-level information in favor of broader trajectory-level features. On the other hand, most representation frameworks~\cite{chang2023contrastive,li2018deep} are restricted to a single form of trajectory data encoding and lack a mechanism to incorporate global context or higher-level information. Recent work~\cite{li2024t} (as shown in Fig.~\ref{fig:intro}, top) explores alternative self-supervised paradigms that capture higher-level semantic information without manual augmentation. Nevertheless, a flexible and semantically aware representation architecture that unifies multiple levels of trajectory information remains an open question. \looseness -1

Sequence models~\cite{vaswani2017attention,hochreiter1997long}, such as recurrent neural networks (RNNs) and Transformers, are a natural choice for trajectory representation due to their ability to process temporally ordered data. However, they exhibit inherent limitations when representing hierarchical semantics of trajectory data. Specifically, these models often operate at a single temporal granularity: they either overemphasize point-level nuances, making them susceptible to noise, or focus too heavily on coarse trajectory-level summaries and thus oversimplify critical details. This single-scale bias in sequential models prevents them from integrating complementary information across abstraction levels and inhibits explicit semantic interactions between local (point-level), intermediate (segment-level), and global (trajectory-level) representations, making it challenging for sequence models to capture long-term dependencies while maintaining the detailed local nuances. Besides, different from uniformly sampled time series data without spatial topology, trajectories are more capricious due to their irregular, geometry-aware, and network-constrained characteristics. 

A new framework is thus required to facilitate the model's understanding of various levels of trajectory representation information, to allow predictions to be grounded on more extensive, multi-dimensional knowledge. In this paper, we propose HiT-JEPA (as shown in Fig.~\ref{fig:intro}, bottom), a hierarchical framework for urban trajectory representation learning, which is designed to address the gaps mentioned above by integrating trajectory semantics across three levels of granularity. Its three-layer architecture explicitly captures (1) point-level details, modeling fine-grained spatial-temporal features of consecutive points; (2) intermediate-level patterns, learning representations of local displacement patterns that reflect mesoscopic movement structures; and (3) high-level abstractions, distilling the overall semantic context as summarized moving behaviors of an entire trajectory. The model unifies multiple information scales within a single representation framework through this hierarchy. Moreover, HiT-JEPA enables interactions between adjacent levels to enrich and align the learned trajectory embeddings across scales. By leveraging a joint embedding predictive architecture, the framework learns to predict and align latent representations between these semantic levels, facilitating semantic integration in a self-supervised manner. For clarity, we summarize our contributions as follows: \looseness -1

\begin{itemize}
    \item We propose HiT-JEPA, a novel hierarchical trajectory representation learning architecture that encapsulates movement information across different semantic levels inside a cohesive framework. HiT-JEPA is the first architecture to explicitly unify both fine-grained and abstract trajectory patterns within a single model.
    \vspace{.7 em}
    \item HiT-JEPA introduces a joint embedding predictive architecture that unifies the entire trajectory across multiple levels of abstraction, resulting in a flexible representation that can seamlessly incorporate local trajectory nuances and global semantic context. By striking a balance between coarse-to-fine trajectory representations by our proposed hierarchical interaction module, we address the limitations of single-scale or single-view models.
    \vspace{.7 em}
    \item We conduct extensive experiments on real-world urban trajectory datasets spanning diverse cities and movement patterns, demonstrating that HiT-JEPA’s semantically enriched, hierarchical embeddings exhibit comparative trajectory similarity search and remarkably superior zero-shot performance across heterogeneous urban and maritime datasets.
\end{itemize}

\section{Related Work} 

\textbf{Urban Trajectory Representation Learning on Similarity Computation.} 
Self-supervised learning methods for trajectory similarity computation are proposed to cope with robust and generalizable trajectory representation learning on large, unlabeled datasets. t2vec~\cite{li2018deep} divides spatial regions into rectangular grids and applies Skip-gram~\cite{mikolov2013efficient} models to convert grid cells into word tokens, then leverages an encoder-decoder framework to learn trajectory representations. TrajCL~\cite{chang2023contrastive} applies contrastive learning on multiple augmentation schemes with a dual-feature attention module to learn both structural and spatial information in trajectories. CLEAR~\cite{li2024clear} proposes a ranked multi-positive contrastive learning method by ordering the similarities of positive trajectories to the anchor trajectories. Recently, T-JEPA~\cite{li2024t} employs a Joint Embedding Predictive Architecture that shifts learning from trajectory data into representation space, establishing a novel self-supervised paradigm for trajectory representation learning. It is also worth noting that robust trajectory representations are often the prerequisite for effective trajectory clustering~\cite{yao2017trajectory,wang2022deep,fang20212}, which focuses on uncovering latent behavioral patterns by grouping trajectories with high semantic affinity.
However, none of the above methods manages to explicitly capture hierarchical trajectory information. We propose HiT-JEPA to support coarse-to-fine, multi-scale trajectory abstraction extraction in a hierarchical JEPA structure.

\noindent\textbf{Hierarchical Self-supervised Learning (HSSL).}
Self-supervised learning methods have significantly advanced the capability to extract knowledge from massive amounts of unlabeled data. Recent approaches emphasize multi-scale feature extraction to achieve a more comprehensive understanding of complex data samples (e.g., lengthy texts or high-resolution images with intricate details). In Computer Vision (CV), Chen \textit{et al.}~\cite{chen2022scaling} stack three Vision Transformers~\cite{dosovitskiy2020image} variants (varying patch size configurations) to learn cell, patch, and region representations of gigapixel whole-slide images in computational pathology. Kong \textit{et al.}~\cite{kong2023understanding} design a hierarchical latent variable model incorporating Masked Autoencoders (MAE)~\cite{he2022masked} to encode and reconstruct multi-level image semantics. Xiao \textit{et al.}~\cite{xiao2022hierarchical} split the hierarchical structure by video semantic levels and employ different learning objectives to capture distinct semantic granularities. In Natural Language Processing (NLP),  Zhang \textit{et al.}~\cite{zhang2019hibert} develop HIBERT, leveraging BERT~\cite{devlin2019bert} to learn sentence-level and document-level text representations for document summarization. Li \textit{et al.}~\cite{li2022hiclre} introduce HiCLRE, a hierarchical contrastive learning framework for distantly supervised relation extraction, utilizing Multi-Granularity Recontextualization for cross-level representation interactions to effectively reduce the influence of noisy data. \looseness -1
In contrast to these methods, which partition inputs into discrete fragments and directly propagate representations across levels, HiT-JEPA encodes the entire trajectory at multiple abstraction levels by coupling adjacent-level attention weights from a hierarchical JEPA to learn multi-scale urban trajectory representations.

\section{Methodology}\label{Methodology}
\begin{figure*}[!ht]
\centering
\includegraphics[width=.8\linewidth]{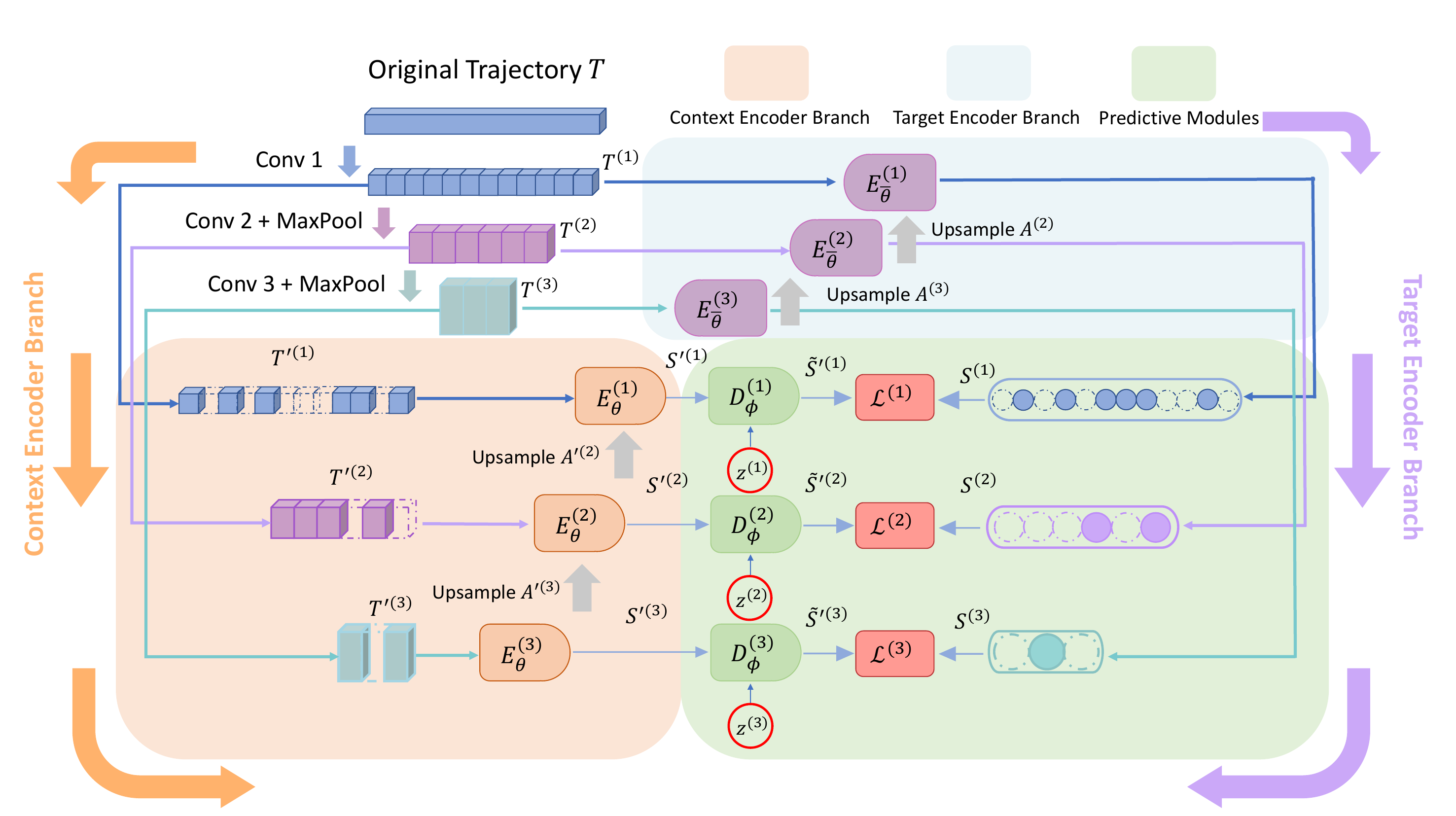}
\setlength{\abovecaptionskip}{-.3 em}
\caption{
HiT-JEPA builds a three-level JEPA hierarchy to extract multi-scale trajectory semantics: (1) Level 1 encodes fine-grained, local point-level features; (2) Level 2 abstracts mesoscopic segment-level patterns; (3) Level 3 captures coarse, global route structures. Trajectory information is propagated from top to bottom, consecutive levels via attention weights.}
\vspace{-1.5 em}
\label{fig:model}
\end{figure*}

Compared to previous methods that only model trajectories at point-level, our primary goal in designing HiT-JEPA is to bridge the gap between simultaneous modeling of local trajectory details and global movement patterns by embedding explicit, cross-level trajectory abstractions into a JEPA framework. To that end, as Fig. \ref{fig:model} illustrates, given a trajectory $T$, we apply three consecutive convolutional layers followed by max pooling operations to produce point-level representation $T^{(1)}$, intermediate-level semantics $T^{(2)}$ and high-level summary $T^{(3)}$, where higher layer representations consist of coarser but semantically richer trajectory patterns. Trajectory abstraction at layer $l$ is learned by the corresponding JEPA layer $\mathrm{JEPA}^{(l)}$ to capture multi-scale sequential dependencies.

\textbf{Spatial region representation.}
Considering the continuous and high-precision nature of GPS coordinates, we partition the continuous spatial regions into fixed cells. But different from previous approaches~\cite{chang2023contrastive,li2024t,li2024clear} that use grid cells, we employ Uber H3\footnote{https://h3geo.org/} to map GPS points into hexagonal grids to select the grid cell resolutions adaptively according to the study area size. Each hexagonal cell shares six equidistant neighbors, with all neighboring centers located at the same distance from the cell’s center. Therefore, we structurally represent the spatial regions by a graph $\mathcal{G}=(V,E)$ where each node $v_{i} \in V$ is a hexagon cell connecting to its neighboring cells $v_{j} \in V$ by an undirected edge $e_{ij} \in E$. We pretrain the spatial node embeddings $\mathcal{H}$ of graph $\mathcal{G}$ using node2vec~\cite{grover2016node2vec}, which produces an embedding set:
\begin{equation}
    \mathcal{H} = \bigl\{\,h_i \in \mathbb{R}^d : v_i \in V\bigr\}, 
\end{equation}
where each $h_{i}$ encodes the relative position of node $v_{i}$. For a GPS location $P=(lon, lat)$, we first assign it to its grid cell index via:
\begin{equation}
    \delta\colon \mathbb{R}^2 \to \{1,\dots,|V|\}, 
\end{equation}
and then look up its embedding $h_{\delta(p)} \in \mathcal{H}$.

\textbf{Hierarchical trajectory abstractions.}
After obtaining the location embeddings, we construct trajectory representations at multiple semantic levels, which are termed hierarchical trajectory abstractions. Given a trajectory $T$ with length $n$, we obtain its location embeddings and denote the input trajectory as $T=(h_{\delta(t_1)}, h_{\delta(t_2)}, \ldots, h_{\delta(t_n)}) \in (\mathbb{R}^{d})^n$. Then, we create its multi-level abstractions $T^{(1)}$, $T^{(2)}$, $T^{(3)}$ by a set of convolutions with kernel size of 3 and stride of 2, and max pooling layers:
\vspace{-.5 em}
{\large
\begin{align}
T^{(1)} &= \mathrm{LayerNorm}(\mathrm{MaxPool1D}(\mathrm{Conv1D}(T))) \notag \\
        &\quad \in (\mathbb{R}^{d})^{n_1}, \quad n_1 = n, \label{eq:l1} \\[6pt] 
T^{(2)} &= \mathrm{LayerNorm}(\mathrm{MaxPool1D}(\mathrm{Conv1D}(T^{(1)}))) \notag \\
        &\quad \in (\mathbb{R}^{d})^{\,n_2}, \quad n_2 = \lfloor n_1/2 \rfloor, \label{eq:l2} \\[6pt]
T^{(3)} &= \mathrm{LayerNorm}(\mathrm{MaxPool1D}(\mathrm{Conv1D}(T^{(2)}))) \notag \\
        &\quad \in (\mathbb{R}^{d})^{\,n_3}, \quad n_3 = \lfloor n_2/2 \rfloor. \label{eq:l3}
\end{align}
}

where $T^{(1)}$ in layer $1$ preserves the channel dimension $d$ and sequence length $n_1=n$, $T^{(2)}$ in layer $2$ keeps the channel dimension and halves the sequence length to $n_2=n/2$, and $T^{(3)}$ in layer $3$ also keeps the channel dimension and halves the sequence length to $n_3=n/4$. Higher-layer trajectory abstractions contain aggregated, high-level trajectory semantic behaviors, while lower layers preserve fine‐grained, local dynamic details.

\textbf{Target encoder branch.}
For the target encoder branch, at each level $l \in \{1,2,3\}$ the target trajectory representation is extracted by:
\begin{equation}
    S^{(l)} = E_{\bar{\theta}}^{(l)}(T^{(l)})
\end{equation}
where $E_\theta^{(l)}$ is the target encoder at layer $l$. Similar to previous JEPA methods~\cite{lecun2022path,assran2023self,li2024t,bardes2023v}, we randomly sample $M$ times from target representation to create the targets, where $S^{(l)}(i)=\{S_{j}^{(l)}\}_{j \in \mathcal{M}_{i}}$. Therefore, $S^{(l)}(i)$ is the $i$-th sampled target and $\mathcal{M}_{i}$ is the $i$-th sampling mask starting from a random position. To ensure the diversity of learning targets, we follow T-JEPA~\cite{li2024t} and introduce a set of masking ratios $r=\{r_1, r_2, r_3, r_4, r_5\}$ where each ratio value specifies the fraction of the representation to mask. At each sampling step, we uniformly draw one ratio from r. We also introduce a probability $p$: with probability $p$, we apply successive masking, and with probability $1-p$, we scatter the masks randomly. Successive masking encourages the encoder to learn both local and long-range dependencies. \looseness -1

\textbf{Context encoder branch.}
For the context encoder branch, we initially sample a trajectory context $C^{(l)}$ from $T^{(l)}$ at level $l$ by a mask $\mathcal{C}_T$ at with sampling ratio $p_{\gamma}$. Next, to prevent any information leakage, we remove from $C^{(l)}$ all positions that overlap with the targets $S^{(l)}$ to obtain the context input $T{'}^{(l)}$. The context trajectory representation $S{'}^{(l)}$ at level $l$ is extracted by:
\begin{equation}
    S{'}^{(l)} = E_\theta^{(l)}(T{'}^{(l)})
\end{equation}

where $E_\theta^{(l)}$ is the context encoder at level $l$. During inference, we use $S{'}^{(1)}$ from $E_\theta^{(1)}$, enriched by the full hierarchy of multi-scale abstractions, as the final output of trajectory representations for similarity comparison or downstream fine-tuning.

\textbf{Predictions.}
Once we have both context representations $S{'}^{(l)}$ and targets $S^{(l)}$ at level $l$, we apply JEPA predictor $D_{\phi}^{(l)}$ on $S{'}^{(l)}$ to approximate $S^{(l)}$ with the help of the mask tokens $z^{(l)}$:
\begin{equation}
    \widetilde{S}{'}^{(l)}(i) = D_{\phi}^{(l)}(\mathrm{CONCAT}(S{'}^{(l)}, \mathrm{PE}(i)\oplus(z^{(l)})))
\end{equation}
where $\mathrm{CONCATE(\cdot)}$ denotes concatenation and $\text{PE}(i)$ refers to the positional embedding after applying the target sampling mask $\mathcal{M}_{i}$. $\oplus$ is element-wise addition between these masked positional embeddings and the mask tokens. Then, we concatenate the mask tokens with positional information with the context representations to guide the predictor in approximating the missing components in the targets at the representation space.

\textbf{Hierarchical interactions.}
By applying JEPA independently at each level, we learn trajectory representations at multiple scales of abstractions. However, the encoders at each level remain siloed and retain only their scale-specific information without leveraging insights from other layers. To enable hierarchical and multi-scale feature extraction, we propagate high-level information down to the next lower abstraction layer.

We adopt Transformer encoders~\cite{vaswani2017attention} for both context and target encoders as their self-attention module is proven highly effective in sequential modeling. Therefore, for both branches, we inject attention weights to the next lower level as a ``top-down spotlight'' where the high-level encoder casts its attention maps to the lower layer, lighting up where the lower‐level encoder should attend. For clarity, we illustrate the process using the target encoder branch as an example. At level $l$, given the query and key matrices $Q^{(l)}$ and $K^{(l)}$ of an input trajectory abstraction $T^{(l)}$, we first retrieve the attention coefficient by:
\begin{align}
    d_k &= \frac{d^{(l)}}{H}, \quad Q_i^{(l)} = Q^{(l)} W_i^{Q,(l)}, \quad K_i^{(l)} = K^{(l)} W_i^{K,(l)}, \label{eq:params} \\
    A_i^{(l)} &= \mathrm{softmax} \left( \frac{Q_i^{(l)} (K_i^{(l)})^\top}{\sqrt{d_k}} \right), \quad i=1,\dots,H \label{eq:attention}
\end{align}
where $H$ is the number of attention heads, $W_i^{Q,(l)}$ and $W_i^{K,(l)}$ are head-$i$ projections, $d^{(l)}$ is the channel dimension, and $A_i^{(l)}$ is the attention coefficient of the head-$i$. The multi-head attention coefficient $A^{(l)}$ are concatenated and projected by:
\begin{equation}
    A^{(l)} = \mathrm{Concat}\bigl(A_1^{(l)},\dots,A_H^{(l)}\bigr)\;W^{O,(l)}
\end{equation}
where $W^{O,(l)}$ is the multi-head projection. To construct the output representation $S^{(l)}$ at level $l$, we simply apply the value matrix $V^{(l)}$ by:
\begin{equation}
    S^{(l)} = A^{(l)}\,V^{(l)}
\end{equation}

Since the dimension of $A^{(l)}$ is:
\begin{equation}
    A^{(l)} \in [0,1]^{\,n^{(l)}\times n^{(l)}},
    \quad n^{(l)} = \Bigl\lfloor\frac{n^{(l-1)}}{2}\Bigr\rfloor
\end{equation}
where $n^{(l)}$ is the length of trajectory abstractions at level $l$, which is half of $n^{(l-1)}$ at level $l-1$ due to Eq.~\ref{eq:l2} and Eq.~\ref{eq:l3}. We need to upsample the attention coefficients:
\begin{equation}
    \widetilde A^{(l)}
    \;=\;
    \mathrm{Interpolate}_{\mathrm{bilinear}}\bigl(A^{(l)}\bigr)
    \;\in\;[0,1]^{\,n^{(l-1)}\times n^{(l-1)}}
\end{equation}
where we adopt bilinear interpolation to upsample the attention weights at level $l$.
To propagate the upsampled $\widetilde A^{(l)}$ to the next lower level, We refer to~\cite{chang2023contrastive} to calculate a weighted sum between $\widetilde A^{(l)}$ and lower level attention coefficient $A^{(l-1)}$. Therefore, we obtain the updated attention coefficient $A^{(l-1)}$ at level $l-1$ by:

\begin{equation}
    A^{(l-1)} = (\sigma A^{(l-1)} + (1-\sigma) \widetilde A^{(l)})
\end{equation}

where $\sigma$ is a learnable scale factor weighting the importance of $A^{(l)}$. Attention coefficient $A{'}^{(l)}$ from the context encoders follows an identical procedure. Global insights guide next-layer feature extraction to focus on semantically important segments. This alignment sharpens local feature extraction, ensuring consistency with the overall context. \looseness -1

\textbf{Loss function.}
After obtaining the predicted representation $\widetilde{S}{'}^{(l)}(i)$ and the $i$-th target representation $S^{(l)}(i)$ at level $l$, we apply $\mathrm{SmoothL1}$ to calculate the loss $\mathcal{L}^{(l)}$ between them:

{\small
\begin{equation}
\begin{aligned}
\mathcal{L}^{(l)} 
&= \underbrace{\frac{1}{M B} \sum_{i=1}^{M}\sum_{b=1}^{B} \sum_{n=1}^{N^{(l)}}\sum_{k=1}^{d^{(l)}} \mathrm{SmoothL1}\bigl(\widetilde S'^{(l)}(i)_{b,n,k}, S^{(l)}(i)_{b,n,k}\bigr)}_{\mathcal{L}_{\mathrm{JEPA}}^{(l)}} \\[6pt]
&\quad + \underbrace{\mathrm{VarLoss}\bigl(z_{\mathrm{tar}}^{(l)}\bigr) + \mathrm{VarLoss}\bigl(z_{\mathrm{ctx}}^{(l)}\bigr) + \mathrm{CovLoss}\bigl(z_{\mathrm{tar}}^{(l)}\bigr) + \mathrm{CovLoss}\bigl(z_{\mathrm{ctx}}^{(l)}\bigr)}_{\mathcal{L}_{\mathrm{VICReg}}^{(l)}}.
\end{aligned}
\end{equation}
}

where we sum over the channel and sequence length dimension $d^{(l)}$ and $N^{(l)}$, and average over the batch and number of target masks dimension $B$ and $M$ to obtain JEPA loss $\mathcal{L}_{\mathrm{JEPA}}^{(l)}$. We also add VICReg~\cite{bardes2021vicreg} to prevent representation collapse, yielding more discriminative representations. We obtain the regularization term $\mathcal{L}_{\mathrm{VICReg}}^{(l)}$ by summing up the variance loss $\mathrm{VarLoss}(\cdot)$ and covariance loss $\mathrm{CovLoss}(\cdot)$ of both expanded context representation $z_{\mathrm{ctx}}^{(l)} = \mathrm{MLP}(S{'}^{(l)})$ and expanded target representation $z_{\mathrm{tar}}^{(l)} = \mathrm{MLP}(S^{(l)})$ via a single-layer MLP. Afterwards, $\mathcal{L}_{\mathrm{VICReg}}^{(l)}$ is added to the loss $\mathcal{L}^{(l)}$ at level $l$.

For level $l \in \{1,2,3\}$, we calculate a weighted sum to obtain the final loss $\mathcal{L}$:
\begin{equation}
    \mathcal{L} = \lambda * \mathcal{L}^{(1)} + \mu * \mathcal{L}^{(2)} + \nu * \mathcal{L}^{(3)} 
\end{equation}
where $\lambda$, $\mu$ and $\nu$ are the scale factors for loss at each level.
\section{Experiments}\label{Experiments}

We conduct experiments on three real-world urban GPS trajectory datasets: Porto \footnote{https://www.kaggle.com/c/pkdd-15-predict-taxi-service-trajectory-i/data}, T-Drive~\cite{yuan2011driving,yuan2010t} and GeoLife~\cite{zheng2008understanding,zheng2009mining,zheng2010geolife}, two FourSquare datasets: FourSquare-TKY and FourSquare-NYC~\cite{yang2014modeling}, and one vessel trajectory dataset: Vessel Tracking Data Australia, which we call  ``AIS(AU)'' \footnote{https://www.operations.amsa.gov.au/spatial/DataServices/DigitalData}.

\begin{table}[!htbp]
\centering
\caption{Statistics of datasets after preprocessing.}
\label{tab:data_summary}
\begin{tabular}{@{}clrr@{}}
\toprule
Data type & Dataset & \#points & \#trajectories \\
\midrule
\multirow{3}{*}{Urban trajectories}
  & Porto   & 65,913,828 & 1,372,725 \\
  & T-Drive & 5,579,067  & 101,842   \\
  & GeoLife & 8,987,488  & 50,693    \\
\cmidrule(l){2-4}
\multirow{2}{*}{Check-in sequences}
  & TKY & 106,480 & 3,048 \\
  & NYC & 28,858  & 734   \\
\cmidrule(l){2-4}
Vessel trajectories & AIS(AU) & 485,424 & 7,095 \\
\bottomrule
\end{tabular}
\end{table}

\subsection{Datasets}\label{datasets}

Here we list the details of the datasets:
\begin{itemize}
    \item \textbf{Porto} includes 1.7 million trajectories from 442 taxis in Porto, Portugal. The dataset was collected from July 2013 to June 2014.
    \item \textbf{T-Drive} contains trajectories of 10,357 taxis in Beijing, China from Feb. 2 to Feb. 8, 2008. The average sampling interval is 3.1 minutes.
    \item \textbf{GeoLife} contains trajectories of 182 users in Beijing, China from April 2007 to August 2012. There are 17,6212 trajectories in total with most of them sampled in 1--5 seconds.
    \item \textbf{Foursquare-TKY} is collected for 11 months from April 2012 to February 2013 in Tokyo, Japan, with 573,703 check-ins in total.
    \item \textbf{Foursquare-NYC} is collected for 11 months from April 2012 to February 2013 in New York City, USA, with 227,428 check-ins in total.
    \item \textbf{AIS(AU)} comprises vessel traffic records collected by the Craft Tracking System (CTS) of Australia. In this paper, we use vessel trajectories in February 2025.
\end{itemize}

\renewcommand{\arraystretch}{0.9}
\setlength{\tabcolsep}{3pt}

\begin{table*}[!tb]
  \centering
  \caption{Mean-rank comparison of methods across meta ratios $R_1$\(\sim\)$R_5$. For each meta ratio, we report the mean ranks under varying DB size $|D|$, downsampling rate $\rho_{s}$, and distortion rate $\rho_{d}$. \textbf{Bold} values are the lowest mean ranks and \underline{underlined} values are the second lowest.}
  \label{tab:main_table}
  \resizebox{\textwidth}{!}{%
    \begin{tabular}{ll*{5}{ccc}}
    \toprule
    Dataset & Method
    & \multicolumn{3}{c}{$R_1$}
    & \multicolumn{3}{c}{$R_2$}
    & \multicolumn{3}{c}{$R_3$}
    & \multicolumn{3}{c}{$R_4$}
    & \multicolumn{3}{c}{$R_5$} \\
    \cmidrule(lr){3-5} \cmidrule(lr){6-8}
    \cmidrule(lr){9-11} \cmidrule(lr){12-14} \cmidrule(lr){15-17}
    & 
    & $|D|$ & $\rho_{s}$ & $\rho_{d}$
    & $|D|$ & $\rho_{s}$ & $\rho_{d}$
    & $|D|$ & $\rho_{s}$ & $\rho_{d}$
    & $|D|$ & $\rho_{s}$ & $\rho_{d}$
    & $|D|$ & $\rho_{s}$ & $\rho_{d}$ \\
    \midrule

    \multirow{10}{*}{Porto}
        & EDR
          &   6.877 &  41.876 &  25.656
          &  12.792 & 122.337 &  25.122
          &  16.052 & 401.732 &  24.494
          &  20.645 & 1127.108 &  23.918
          &  26.101 & 2293.710 &  23.993 \\
        & LCSS
          &  52.502 & 224.574 & 230.361
          & 103.628 & 253.337 & 229.337
          & 137.337 & 311.083 & 227.792
          & 181.325 & 385.245 & 225.323
          & 231.807 & 492.691 & 224.426 \\
        & Hausdorff
          &   2.597 &   7.531 &   7.890
          &   3.960 &  13.256 &   7.718
          &   4.914 &  11.809 &   8.713
          &   6.093 &  16.237 &   8.570
          &   7.518 &  28.808 &   6.848 \\
        & Fr\'{e}chet
          &   4.162 &  13.856 &  14.095
          &   7.344 &  16.166 &  13.667
          &   9.144 &  31.097 &  15.129
          &  11.562 &  43.857 &  14.186
          &  14.472 &  74.390 &  12.592 \\
    \cmidrule(lr){2-17}
      & TrajCL
        & \textbf{1.004} & \textbf{1.047} & \textbf{1.017}
        & \textbf{1.007} & \textbf{1.170} & \textbf{1.029}
        & \textbf{1.008} & \textbf{1.905} & \textbf{1.036}
        & \textbf{1.011} & \textbf{6.529} & \textbf{1.060}
        & \textbf{1.014} & 68.557 & \textbf{1.022} \\
      & TrjSR
        & 3.240 & 12.553 & 12.509
        & 5.321 & 16.945 & 15.401
        & 7.073 & 37.150 & 15.901
        & 8.740 & 65.413 & 28.914
        & 10.192 & 149.950 & 32.730 \\
      & CLEAR
        & 3.235 & 7.796  & 4.250
        & 4.012 & 13.323 & 4.442
        & 4.088 & 22.814 & 4.284
        & 4.137 & 44.865 & 4.438
        & 4.204 & 123.921 & 4.399 \\
      & T-JEPA
        & 1.029 & 1.455 & 1.097
        & 1.048 & \underline{2.304} & 1.084
        & 1.053 & \underline{4.413} & 1.115
        & 1.061 & \underline{9.599} & 1.110
        & 1.074 & \textbf{23.900} & 1.123 \\
      & BLUE
        & 2.576 & 18.956 & 9.587
        & 3.707 & 31.181 & 17.926
        & 4.834 & 54.553 & 16.638
        & 5.914 & 111.849 & 32.330
        & 6.952 & 250.886 & 41.010 \\
      & HiT-JEPA
        & \underline{1.026} & \underline{1.369} & \underline{1.074}
        & \underline{1.043} & 2.624 & \underline{1.077}
        & \underline{1.048} & 5.541 & \underline{1.085}
        & \underline{1.058} & 13.773 & \underline{1.093}
        & \underline{1.065} & \underline{28.806} & \underline{1.119} \\
    \midrule

    \multirow{10}{*}{T-Drive}
        & EDR
          &  47.461 & 313.419 & 222.460
          &  90.016 & 558.132 & 221.547
          & 135.747 & 830.649 & 224.160
          & 181.244 & 1136.547 & 222.132
          & 223.593 & 1422.003 & 224.402 \\
        & LCSS
          &   5.697 &  21.462 &  20.816
          &   8.392 &  24.296 &  20.762
          &  13.325 &  29.944 &  20.698
          &  17.180 &  36.488 &  20.617
          &  20.874 &  53.564 &  20.538 \\
        & Hausdorff
          &  13.165 &  61.861 &  59.387
          &  24.435 &  51.330 &  59.438
          &  35.985 &  33.801 &  59.559
          &  47.927 &  56.741 &  59.346
          &  59.487 &  61.825 &  59.281 \\
        & Fr\'{e}chet
          &  11.901 &  53.559 &  52.624
          &  21.741 &  45.354 &  52.586
          &  31.992 &  27.896 &  52.632
          &  42.557 &  41.717 &  52.549
          &  52.675 &  45.183 &  52.519 \\
    \cmidrule(lr){2-17}
      & TrajCL
        & 1.111 & 1.203 & 1.267
        & 1.128 & 1.348 & 3.320
        & 1.146 & 1.668 & 1.355
        & 1.177 & \underline{1.936} & 1.513
        & 1.201 & \underline{3.356} & 1.179 \\
      & TrjSR
        & 110.726 & 674.16 & 581.776
        & 223.841 & 795.331 & 572.944
        & 356.941 & 870.73 & 566.816
        & 475.872 & 960.404 & 545.278
        & 592.146 & 1033.404 & 566.696 \\
      & CLEAR
        & 1.047 & 1.305 & 1.111
        & 1.062 & 1.484 & 1.110
        & 1.077 & 1.964 & 1.171
        & 1.088 & 3.497 & 1.152
        & 1.104 & 3.902 & 1.172 \\
      & T-JEPA
        & \textbf{1.032} & \underline{1.088} & \underline{1.054}
        & \textbf{1.034} & \underline{1.225} & \underline{1.061}
        & \textbf{1.036} & \underline{1.617} & \underline{1.069}
        & \underline{1.045} & 3.226 & \underline{1.067}
        & \underline{1.049} & 4.115 & \underline{1.078} \\
      & BLUE
        & 10.707 & 48.475 & 47.958
        & 18.735 & 64.435 & 48.448
        & 26.688 & 102.816 & 37.423
        & 34.592 & 108.631 & 37.838
        & 42.587 & 150.521 & 34.917 \\
      & HiT-JEPA
        & \underline{1.040} & \textbf{1.057} & \textbf{1.035}
        & \underline{1.040} & \textbf{1.085} & \textbf{1.029}
        & \underline{1.040} & \textbf{1.172} & \textbf{1.039}
        & \textbf{1.040} & \textbf{1.389} & \textbf{1.033}
        & \textbf{1.040} & \textbf{2.222} & \textbf{1.034} \\
    \midrule

    \multirow{10}{*}{GeoLife}
        & EDR
          &  64.467 & 334.605 & 333.229
          & 129.798 & 342.666 & 327.207
          & 162.822 & 341.532 & 322.759
          & 306.303 & 359.849 & 320.356
          & 339.344 & 345.723 & 317.877 \\
        & LCSS
          &  70.498 & 374.889 & 369.087
          & 143.521 & 375.423 & 364.732
          & 181.778 & 373.907 & 361.717
          & 335.844 & 376.130 & 360.432
          & 373.353 & 378.057 & 358.623 \\
        & Hausdorff
          &   3.166 &   8.281 &   9.800
          &   4.341 &   6.277 &   9.563
          &   5.051 &   8.630 &   9.582
          &   7.746 &   8.656 &   8.978
          &   8.263 &   8.343 &   9.438 \\
        & Fr\'{e}chet
          &   3.443 &   9.294 &  10.432
          &   4.687 &   7.408 &  10.423
          &   5.425 &  10.007 &  10.423
          &   8.468 &   9.647 &  10.703
          &   9.172 &   9.728 &  10.107 \\
    \cmidrule(lr){2-17}
      & TrajCL
        & 1.130 & 1.440 & 7.973
        & 1.168 & 1.435 & 19.266
        & 1.195 & 1.720 & 12.397
        & 1.234 & 1.616 & 10.560
        & 1.256 & 2.675 & 11.035 \\
      & TrjSR
        & 6.765 & 8.332 & 7.747
        & 7.393 & 8.594 & 7.942
        & 7.661 & 8.688 & 7.648
        & 7.767 & 8.566 & 8.534
        & 8.350 & 8.770 & 9.460 \\
      & CLEAR
        & 1.110 & 1.196 & 1.212
        & 1.124 & 1.318 & \underline{1.211}
        & 1.144 & 1.818 & 1.189
        & 1.145 & 2.237 & 1.239
        & 1.155 & 3.712 & 1.333 \\
      & T-JEPA
        & \textbf{1.019} & \textbf{1.052} & \textbf{1.047}
        & \underline{1.034} & \textbf{1.030} & \textbf{1.093}
        & \underline{1.036} & \textbf{1.103} & \textbf{1.101}
        & \underline{1.040} & \textbf{1.150} & \textbf{1.154}
        & \underline{1.047} & \textbf{1.218} & \textbf{1.197} \\
      & BLUE
        & 1.445 & 2.808 & 25.613
        & 1.720 & 4.025 & 30.134
        & 1.924 & 6.840 & 32.967
        & 2.122 & 4.887 & 23.083
        & 2.377 & 4.386 & 27.894 \\
      & HiT-JEPA
        & \underline{1.033} & \underline{1.061} & \underline{1.170}
        & \textbf{1.033} & \underline{1.111} & \underline{1.370}
        & \textbf{1.033} & \underline{1.247} & \underline{1.357}
        & \textbf{1.033} & \underline{1.377} & \underline{1.509}
        & \textbf{1.033} & \underline{1.573} & \underline{1.511} \\
    \midrule

    \multirow{10}{*}{\shortstack{TKY\\(zero-shot)}}
        & EDR
          &  72.453 & 371.982 & 337.370
          & 143.372 & 450.190 & 338.495
          & 216.370 & 527.870 & 335.778
          & 276.345 & 596.542 & 336.118
          & 338.352 & 637.338 & 337.973 \\
        & LCSS
          &   8.403 &  25.565 &  22.773
          &  13.443 &  36.448 &  22.815
          &  16.998 &  47.748 &  22.720
          &  18.960 &  63.605 &  22.685
          &  22.863 &  83.547 &  22.758 \\
        & Hausdorff
          &  24.560 & 113.567 & 110.703
          &  46.892 & 128.057 & 110.567
          &  67.725 & 129.387 & 110.453
          &  89.328 & 168.717 & 109.778
          & 110.728 & 191.167 & 110.470 \\
        & Fr\'{e}chet
          &  63.988 & 332.987 & 327.215
          & 132.443 & 357.870 & 325.568
          & 193.948 & 364.610 & 326.060
          & 259.787 & 395.808 & 325.560
          & 326.818 & 404.967 & 325.135 \\
    \cmidrule(lr){2-17}
      & TrajCL
        & 17.590 & 66.963 & 75.397
        & 32.377 & 67.835 & 79.228
        & 46.958 & 116.677 & 59.222
        & 62.145 & 170.460 & 69.642
        & 78.722 & 211.487 & 65.258 \\
      & TrjSR
        & 8.673 & 31.770 & 27.505
        & 17.120 & 37.070 & 30.758
        & 22.310 & 48.985 & 30.923
        & 26.820 & 64.380 & 33.113
        & 29.318 & 84.302 & 34.043 \\
      & CLEAR
        & 119.561 & 591.345 & 583.863
        & 242.493 & 626.075 & 591.460
        & 349.132 & 646.160 & 587.138
        & 456.525 & 662.553 & 588.212
        & 577.238 & 709.903 & 591.107 \\
      & T-JEPA
        & \underline{1.948} & \underline{3.060} & \underline{3.245}
        & \underline{2.272} & \underline{4.227} & \underline{3.165}
        & \underline{2.617} & \underline{7.975} & \underline{3.313}
        & \underline{2.913} & \underline{18.173} & \underline{3.202}
        & \underline{3.275} & \underline{19.135} & \underline{3.127} \\
      & BLUE
        & 49.847 & 135.002 & 118.957
        & 67.890 & 155.228 & 122.083
        & 84.853 & 181.575 & 116.502
        & 101.950 & 208.765 & 117.895
        & 119.862 & 293.207 & 116.375 \\
      & HiT-JEPA
        & \textbf{1.508} & \textbf{2.490} & \textbf{2.060}
        & \textbf{1.707} & \textbf{2.962} & \textbf{2.002}
        & \textbf{1.835} & \textbf{4.985} & \textbf{2.067}
        & \textbf{1.930} & \textbf{10.268} & \textbf{2.045}
        & \textbf{2.057} & \textbf{14.755} & \textbf{1.988} \\
    \midrule

    \multirow{10}{*}{\shortstack{NYC\\(zero-shot)}}
        & EDR
          &  19.086 &  82.371 &  76.907
          &  35.957 &  92.129 &  76.943
          &  51.079 &  94.529 &  77.271
          &  64.257 & 105.657 &  77.950
          &  77.014 & 107.900 &  76.071 \\
        & LCSS
          &   9.164 &  21.343 &  19.393
          &  10.779 &  23.986 &  19.329
          &  13.807 &  21.607 &  19.150
          &  17.014 &  27.086 &  19.414
          &  19.279 &  28.521 &  19.250 \\
        & Hausdorff
          &   7.014 &  31.436 &  26.621
          &  11.921 &  33.207 &  26.536
          &  17.329 &  40.057 &  26.521
          &  21.707 &  40.457 &  26.750
          &  26.650 &  57.100 &  26.636 \\
        & Fr\'{e}chet
          &  15.350 &  78.471 &  80.357
          &  33.479 &  83.707 &  79.879
          &  50.379 &  83.743 &  79.757
          &  63.871 &  91.836 &  80.293
          &  80.057 &  92.064 &  79.921 \\
    \cmidrule(lr){2-17}
      & TrajCL
        & 4.336 & 16.886 & 15.093
        & 6.457 & 18.857 & 16.971
        & 9.129 & 22.007 & 16.443
        & 12.350 & 37.579 & 11.236
        & 15.071 & 36.650 & 6.543 \\
      & TrjSR
        & 3.929 & 5.457 & 6.307
        & 4.793 & 5.171 & 7.950
        & 5.457 & 8.350 & 6.679
        & 5.821 & 12.757 & 7.443
        & 6.007 & 14.329 & 7.907 \\
      & CLEAR
        & 19.693 & 68.843 & 68.057
        & 32.171 & 74.964 & 68.321
        & 43.214 & 75.121 & 69.221
        & 55.507 & 79.514 & 70.507
        & 67.207 & 84.421 & 65.914 \\
      & T-JEPA
        & \underline{1.450} & \underline{1.950} & \underline{1.714}
        & \underline{1.514} & \underline{3.050} & \underline{1.736}
        & \underline{1.571} & \underline{2.400} & \underline{1.679}
        & \underline{1.636} & \underline{2.457} & \underline{1.771}
        & \underline{1.714} & \underline{5.850} & \underline{1.807} \\
      & BLUE
        & 12.45 & 46.536 & 44.386
        & 20.350 & 60.164 & 43.321
        & 27.907 & 78.136 & 41.329
        & 35.664 & 76.500 & 42.564
        & 42.871 & 86.493 & 42.357 \\
      & HiT-JEPA
        & \textbf{1.343} & \textbf{1.743} & \textbf{1.493}
        & \textbf{1.364} & \textbf{2.143} & \textbf{1.500}
        & \textbf{1.414} & \textbf{1.636} & \textbf{1.500}
        & \textbf{1.457} & \textbf{2.407} & \textbf{1.550}
        & \textbf{1.500} & \textbf{3.343} & \textbf{1.471} \\
    \midrule

    \multirow{10}{*}{\shortstack{AIS(AU)\\(zero-shot)}}
        & EDR
          &  62.017 & 208.320 & 210.012
          & 118.344 & 206.546 & 210.011
          & 157.764 & 204.515 & 210.004
          & 184.769 & 204.077 & 209.993
          & 210.016 & 223.721 & 210.001 \\
        & LCSS
          &  95.162 & 308.838 & 312.060
          & 177.483 & 305.615 & 312.060
          & 234.421 & 303.067 & 312.056
          & 274.775 & 297.921 & 312.040
          & 312.060 & 294.649 & 312.047 \\
        & Hausdorff
          &  20.862 &  64.504 &  64.134
          &  36.845 &  64.474 &  64.239
          &  48.582 &  64.981 &  64.170
          &  57.062 &  65.661 &  64.138
          &  64.099 &  67.696 &  63.943 \\
        & Fr\'{e}chet
          &  13.734 &  50.794 &  49.529
          &  25.926 &  51.786 &  49.606
          &  35.541 &  54.110 &  49.618
          &  43.260 &  57.427 &  49.630
          &  49.492 &  63.237 &  49.493 \\
    \cmidrule(lr){2-17}
      & TrajCL
        & 9.057 & 37.721 & 37.866
        & 18.771 & 9.878 & 37.879
        & 26.538 & 41.068 & 37.862
        & 33.004 & 45.352 & 37.911
        & 37.866 & 48.651 & 38.399 \\
      & TrjSR
        & 692.000 & 3658.400 & 3649.450
        & 1390.364 & 3661.421 & 3649.407
        & 2136.271 & 3675.043 & 3649.150
        & 2942.586 & 3714.564 & 3649.086
        & 2892.264 & 3700.221 & 3649.371 \\
      & CLEAR
        & 38.042 & 188.171 & 184.600
        & 73.164 & 187.914 & 184.579
        & 112.371 & 192.571 & 184.600
        & 150.050 & 191.629 & 184.871
        & 184.600 & 198.843 & 184.593 \\
      & T-JEPA
        & \underline{2.156} & \underline{5.661} & \underline{4.753}
        & \underline{3.176} & \underline{6.849} & \underline{4.753}
        & \underline{3.889} & \textbf{9.486} & \underline{4.755}
        & \underline{4.364} & \textbf{13.055} & \underline{4.758}
        & \underline{4.754} & \textbf{16.986} & \underline{4.749} \\
      & BLUE
        & 21.998 & 67.701 & 77.694
        & 34.434 & 96.408 & 74.085
        & 47.271 & 115.352 & 72.941
        & 59.799 & 169.311 & 73.140
        & 72.233 & 208.531 & 71.264 \\
      & HiT-JEPA
        & \textbf{1.483} & \textbf{4.119} & \textbf{2.759}
        & \textbf{1.954} & \textbf{6.357} & \textbf{2.759}
        & \textbf{2.311} & \underline{10.233} & \textbf{2.758}
        & \textbf{2.579} & \underline{15.180} & \textbf{2.757}
        & \textbf{2.758} & \underline{20.267} & \textbf{2.755} \\
    \bottomrule
  \end{tabular}%
  }
  \vspace{-1.5 em}
\end{table*}

We first keep trajectories in urban areas with the number of points ranging from 20 to 200, where the statistics of the datasets after preprocessing are shown in Table \ref{tab:data_summary}. We use 200,000 trajectories for Porto, 70,000 for T-Drive, and 35000 for GeoLife as training sets. Each dataset has 10\% of data used for validation. As there are many fewer trajectories in TKY, NYC, and AIS(AU), we use all trajectories in these datasets for testing. For the testing set, we select 100,000 trajectories for Porto, 10,000 for T-Drive and GeoLife, 3000 for TKY, 700 for NYC, and 7000 for AIS(AU). For the downstream fine-tuning task, we select 10,000 trajectories for Porto and T-Drive, and 5000 for GeoLife, where the selected trajectories are split by 7:1:2 for training, validation, and testing. We train Hit-JEPA and all baselines from scratch for Porto, T-Drive, and GeoLife datasets. Then, we load the pre-trained weights from Porto and conduct zero-shot self-similarity experiments on each of the TKY, NYC, and AIS(AU) to evaluate the generalization ability of all models. 

\subsection{Implementation Details}
We use Adam Optimizer for training and optimizing the model parameters across all levels, except for the target encoders. The target encoder at each level $l$ updates its parameters via the exponential moving average of the parameters of the context encoder at the same level. The maximum number of training epochs is 20, and the learning rate is 0.0001, decaying by half every 5 epochs. The embedding dimension $d$ is 256, and the batch size is 64. We apply 1-layer Transformer Encoders for both context and target encoders at each level, with the number of attention heads set to 8 and hidden layer dimension to 1024. We use a 1-layer Transformer Decoder as the predictor at each level $l$ with the number of attention heads set to 8. We use learnable positional encoding for all the encoders and decoders. We set the resampling masking ratio to be selected from $r=\{10\%, 15\%, 20\%, 25\%, 30\%\}$ and the number of sampled targets $M$ to 4 for each trajectory at each model level $l$. The successive sampling probability $p$ is set to 50\%, and the initial context sampling ratio $p_{\gamma}$ is set to range from 85\% to 100\%. The scale factors for the final loss are $\lambda=0.05$, $\mu=0.15$, and $\nu=0.8$. We use a hexagonal cell resolution of 11 for Porto, resolution 10 for T-Drive, GeoLife, TKY, and NYC, and resolution 4 for AIS(AU). 
All experiments are conducted on servers with Nvidia A5000 GPUs.

\subsection{Baselines}
We compare HiT-JEPA with four heuristic trajectory similarity measures: EDR, LCSS, Hausdorff, and Discrete Fréchet, and five most recent self-supervised free space trajectory similarity computation methods: TrajCL~\cite{chang2023contrastive}, TrjSR~\cite{cao2021accurate}, CLEAR~\cite{li2024clear}, T-JEPA~\cite{li2024t}, and BLUE~\cite{zhou2025blurred}. TrajCL is a contrastive learning method that adopts a dual-feature attention module to capture the trajectory details, which has achieved impactful performance on trajectory similarity computation in multiple datasets and experimental settings. TrjSR is a generative model that converts trajectories into gray-scale images. This method reconstructs the high-resolution trajectory image from the low-resolution image by leveraging single-image super-resolution to learn better spatial trajectory representations. CLEAR improves the contrastive learning process by ranking the positive trajectory samples based on their similarities to anchor samples, capturing detailed differences from similar trajectories. T-JEPA utilizes Joint Embedding Predictive Architecture to encode and predict trajectory information in the representation space, which effectively captures necessary trajectory information. BLUE proposes a hierarchical patch-based trajectory representation that progressively reduces GPS coordinate precision to capture both fine-grained spatial-temporal details and high-level travel semantics via a pyramid-structured encoder-decoder Transformer. We run these models from their open-source code repositories with default parameters. \looseness -1 

\subsection{Quantitative Evaluation}
In this section, we evaluate HiT-JEPA and compare it to baselines in three experiments: most similar trajectory search, robustness of learn representations, and generalization with downstream fine-tuning. We combine the first two experiments as ``Self-similarity''.

\subsubsection{Self-similarity}
Following similar experimental settings of previous work~\cite{chang2023contrastive,li2024t}, we construct a Query trajectory set $Q$ and a database trajectory $D$ for the testing set given a trajectory. $Q$ has 1,000 trajectories for Porto, T-Drive, and GeoLife, 600 for TKY, 140 for NYC, and 1400 for AIS(AU). And $D$ has 100,000 trajectories for Porto, 10,000 for T-Drive and Geolife, 3000 for TKY, 700 for NYC, and 7000 for AIS(AU). Details can be found in Appendices \ref{exp}.
Table \ref{tab:main_table} reports the mean ranks of all methods. HiT-JEPA achieves the overall lowest mean ranks on five of the six datasets. Among the urban GPS datasets (Porto, T-Drive, and GeoLife), it attains its lowest ranks on T-Drive. For example, the mean ranks for DB size $\mathrm{|}D\mathrm{|}$ across $20\%$\(\sim\)$100\%$ and for distortion rate $\rho_{d}$ across $0.1$\(\sim\)$0.5$ remain remarkably steady ($1.040$\(\sim\)$1.041$ and $1.031$\(\sim\)$1.038$, respectively). T-Drive contains taxi trajectories with much longer and more irregular sampling intervals ($3.1$ minutes on average). By leveraging a hierarchical structure to capture global, high-level trajectory abstractions, HiT-JEPA learns features that remain invariant to noise and sparse sampling, yielding more robust and accurate representations for sparsely and irregularly sampled trajectories even with limited training data. On GeoLife, we obtain mean ranks comparable to T-JEPA (only $2.8\%$ higher), and we rank second overall on Porto. This is because Porto trajectories lie in an especially dense spatial region, allowing TrajCL to exploit auxiliary cues such as movement speed and orientation to tease apart nearly identical paths. However, reliance on these cues undermines generalization to lower-quality trajectories (e.g., T-Drive) and knowledge transfer to other cities. \looseness -1

Next, we evaluate zero-shot performance on TKY, NYC, and AIS(AU). HiT-JEPA consistently achieves the lowest mean ranks across all database sizes, downsampling rates, and distortion rates. Both TKY and NYC consist of highly sparse and coarse check-in sequences that lack dense trajectory waypoints, which challenges the models' summarization ability. Benefiting from its hierarchical structure, HiT-JEPA first summarizes mobility patterns at a coarse level and then refines the check-in details at finer levels. Crucially, this summarization knowledge is transferred from the dense urban trajectories in Porto, demonstrating that HiT-JEPA learns more generalizable representations than TrajCL by capturing more essential spatiotemporal information from trajectories. Even on AIS(AU), whose trajectories span ocean-wide scales, HiT-JEPA still maintains the lowest overall mean ranks, demonstrating its ability to handle diverse trajectory types that spread across various regional scales. By contrast, although CLEAR outperforms TrajCL on T-Drive and GeoLife, it generalizes poorly in the zero-shot experiments on TKY, NYC, and AIS(AU); BLUE behaves similarly, and despite being the most training-efficient method (Table \ref{tab:effi}), it is markedly less robust to downsampling and distortion and transfers poorly across domains. TrjSR exhibits the weakest overall performance across all datasets, because image-based representations struggle to distinguish fine-grained trajectory differences, a difficulty further exacerbated by lower data quality (e.g., T-Drive). Being training-free, the non-learned measures (EDR, LCSS, Hausdorff, and Fr\'{e}chet) apply the same fixed geometric rules to every dataset and trail all learned methods by a large margin. Their behavior is also inconsistent across datasets: Hausdorff and Fr\'{e}chet are competitive only on the dense, regularly sampled Porto and GeoLife, whereas EDR and LCSS degrade severely. Lacking any learned invariance, they cannot adapt to downsampling or distortion and cannot exploit dataset-specific structure, which leads to consistently high ranks across all six datasets.

\subsubsection{Downstream Fine-tuning}
To evaluate the generalization ability of HiT-JPEA, we conduct downstream fine-tuning on its learned representations. Specifically, we retrieve and freeze the encoder of HiT-JEPA and other baselines, concatenated with a 2-layer MLP decoder, then train the decoder to approximate the computed trajectory similarities by heuristic approaches. This setting is first proposed by TrajCL~\cite{chang2023contrastive}, then followed by T-JEPA~\cite{li2024t}, to quantitatively assess whether the learned representations can generalize to approach the computational processes underlying each heuristic measure. In real applications, fine-tuned models can act as efficient, ``fast'' approximations of traditional heuristic measures, alleviating their quadratic time-complexity bottleneck. We report hit ratios $\mathrm{HR}@5$ and $\mathrm{HR}@20$ to evaluate the correct matches between top-5 predictions and each of the top-5 and top-20 ground truths. We also report the recall $\mathrm{R}5@20$ to evaluate the correct matches of 
top-5 ground truths from predicted top-20 predictions. We approximate all model representations to 4 heuristic measures: EDR, LCSS, Hausdorff, and Discrete Fréchet. We do not include TrjSR here as its results are proven to be less competitive in ~\cite{chang2023contrastive}.


From Table~\ref{tab:finetune}, HiT-JEPA achieves the highest overall performance. The ``Average'' column reports the mean over all twelve metrics for each model on each dataset. HiT-JEPA outperforms T-JEPA, the strongest baseline, by $21.4\%$ on T-Drive and $7.2\%$ on GeoLife, while essentially matching it on Porto (within $0.2\%$). On T-Drive, HiT-JEPA surpasses T-JEPA on all four distance measures, with the largest gains on Hausdorff and discrete Fr\'{e}chet (relative average improvements of $38.8\%$ and $31.7\%$, respectively), indicating that the hierarchical design is especially beneficial for the sparse and irregularly sampled taxi trajectories in T-Drive. On GeoLife, HiT-JEPA exceeds T-JEPA on every per-measure average---by $5.1\%$, $3.0\%$, $5.2\%$, and $14.7\%$ on EDR, LCSS, Hausdorff, and Fr\'{e}chet---and trails only on a few individual sub-metrics (e.g., HR@5 under EDR and Fr\'{e}chet). On Porto, HiT-JEPA is on par with T-JEPA overall ($0.508$ vs.\ $0.509$), outperforming it on EDR, LCSS, and Fr\'{e}chet ($+1.9\%$, $+1.1\%$, and $+4.4\%$ on average) and trailing only on the Hausdorff measure ($-6.3\%$).

\begin{table*}
\centering

\tiny
\caption{Comparisons with fine-tuning 2-layer MLP decoder. \textbf{Bold} values are the lowest mean ranks and \underline{underlined} values are the second lowest.}
\vspace{-1.em}
  \resizebox{\textwidth}{!}{%
\begin{tabular}{l@{\hspace{0.1em}} | l@{\hspace{0.1em}} | c@{\hspace{0.3em}}  c@{\hspace{0.3em}}  c@{\hspace{0.3em}} | c@{\hspace{0.3em}}  c@{\hspace{0.3em}}  c@{\hspace{0.3em}} | c@{\hspace{0.3em}}  c@{\hspace{0.3em}}  c@{\hspace{0.3em}} |   c@{\hspace{0.3em}}  c@{\hspace{0.3em}} c@{\hspace{0.3em}}| c@{\hspace{0.2em}}}
 \hline
 \multirow{2}*{Dataset} & 
 \multirow{2}*{Method} &
 \multicolumn{3}{c|}{EDR} & 
 \multicolumn{3}{c|}{LCSS} &
 \multicolumn{3}{c|}{Hausdorff} &
 \multicolumn{3}{c|}{Fréchet} &
 \multirow{2}*{Average} \\
 \cline{3-14} & & HR@5$\uparrow$ & HR@20$\uparrow$ & R5@20$\uparrow$ & HR@5$\uparrow$ & HR@20$\uparrow$ & R5@20$\uparrow$ & HR@5$\uparrow$ & HR@20$\uparrow$ & R5@20$\uparrow$ & HR@5$\uparrow$ & HR@20$\uparrow$ & R5@20$\uparrow$ &  \\
 \hline\hline
\multirow{6}*{Porto} 
& TrajCL & 0.137 & 0.179 & 0.301 & 0.329 & 0.508 & 0.663 & 0.456 & 0.574 & 0.803 & 0.412 & 0.526 & 0.734 & 0.468 \\
& TrjSR & 0.085 & 0.083 & 0.157 & 0.162 & 0.197 & 0.292 & 0.166 & 0.192 & 0.304 & 0.157 & 0.173 & 0.288 & 0.188 \\
& CLEAR & 0.078 & 0.075 & 0.142 & 0.164 & 0.198 & 0.293 & 0.152 & 0.131 & 0.232 & 0.192 & 0.165 & 0.316 & 0.178 \\
& T-JEPA & \underline{0.154} & \underline{0.194} & \underline{0.336} & \underline{0.365} & \underline{0.551} & \underline{0.713} & \textbf{0.525} & \textbf{0.633} & \textbf{0.869} & \underline{0.433} & \underline{0.565} & \underline{0.771} & \textbf{0.509} \\
& BLUE & 0.077 & 0.119 & 0.175 & 0.197 & 0.354 & 0.461 & 0.335 & 0.477 & 0.668 & 0.384 & 0.538 & 0.741 & 0.377 \\
& HiT-JEPA & \textbf{0.163} & \textbf{0.197} & \textbf{0.337} & \textbf{0.369} & \textbf{0.558} & \textbf{0.720} & \underline{0.466} & \underline{0.599} & \underline{0.835} & \textbf{0.450} & \textbf{0.587} & \textbf{0.810} & \underline{0.508} \\
\hline
\multirow{6}*{T-Drive} 
& TrajCL & \underline{0.094} & 0.131 & 0.191 & 0.159 & 0.289 & 0.366 & 0.173 & 0.256 & 0.356 & 0.138 & 0.187 & 0.274 & 0.218 \\
& TrjSR & 0.076 & 0.068 & 0.114 & 0.076 & 0.080 & 0.118 & 0.095 & 0.090 & 0.143 & 0.098 & 0.094 & 0.145 & 0.100 \\
& CLEAR & 0.093 & 0.084 & 0.143 & 0.126 & 0.166 & 0.216 & 0.142 & 0.158 & 0.243 & 0.135 & 0.170 & 0.283 & 0.163 \\
& T-JEPA & \underline{0.094} & \underline{0.147} & \underline{0.215} & \underline{0.205} & \underline{0.366} & \underline{0.469} & 0.158 & 0.229 & 0.329 & 0.125 & 0.159 & 0.249 & \underline{0.229} \\
& BLUE & 0.021 & 0.067 & 0.074 & 0.051 & 0.141 & 0.171 & \underline{0.183} & \underline{0.312} & \underline{0.387} & \textbf{0.197} & \textbf{0.311} & \textbf{0.404} & 0.193 \\
& HiT-JEPA & \textbf{0.112} & \textbf{0.170} & \textbf{0.260} & \textbf{0.221} & \textbf{0.384} & \textbf{0.493} & \textbf{0.222} & \textbf{0.316} & \textbf{0.456} & \underline{0.158} & \underline{0.219} & \underline{0.325} & \textbf{0.278} \\
\hline
\multirow{6}*{GeoLife} 
& TrajCL & \underline{0.193} & 0.363 & 0.512 & 0.232 & 0.484 & 0.584 & 0.479 & 0.536 & 0.745 & 0.398 & 0.463 & 0.708 & 0.475 \\
& TrjSR & 0.138 & 0.246 & 0.443 & 0.229 & 0.330 & 0.479 & 0.492 & 0.439 & 0.692 & 0.383 & 0.362 & 0.614 & 0.404 \\
& CLEAR & 0.175 & 0.164 & 0.311 & 0.224 & 0.224 & 0.342 & 0.347 & 0.308 & 0.499 & 0.397 & 0.273 & 0.539 & 0.320 \\
& T-JEPA & \textbf{0.195} & \underline{0.383} & \underline{0.527} & \underline{0.242} & \underline{0.515} & \underline{0.586} & \underline{0.606} & \underline{0.656} & \underline{0.857} & \textbf{0.488} & 0.406 & \underline{0.731} & \underline{0.516} \\
& BLUE & 0.143 & 0.272 & 0.358 & 0.226 & 0.470 & 0.480 & 0.395 & 0.647 & 0.630 & 0.459 & \textbf{0.647} & 0.674 & 0.450 \\
& HiT-JEPA & 0.183 & \textbf{0.414} & \textbf{0.564} & \textbf{0.250} & \textbf{0.525} & \textbf{0.609} & \textbf{0.643} & \textbf{0.700} & \textbf{0.885} & \underline{0.467} & \underline{0.555} & \textbf{0.842} & \textbf{0.553} \\
\hline
\end{tabular}
}
\vspace{-.5em}
\label{tab:finetune}

\end{table*}

\subsection{Training Efficiency}\label{efficiency}

We compare HiT-JEPA with baselines in terms of training time per iteration in Table \ref{tab:effi}. Although HiT-JEPA is not among the fastest methods, its per-iteration time (0.341 s) stays close to other JEPA-based baselines and avoids the heavy cost of TrjSR, indicating that the hierarchical structure does not impose a substantial efficiency penalty. Moreover, by incorporating convolution-based trajectory semantics aggregation and learning on multi-level trajectory abstractions with 1-layer Transformer backbones, HiT-JEPA remains efficient while achieving generalizable and robust performance.

\begin{table}[t]
  \centering
  \caption{Comparison of training efficiency (seconds per iter.).}
  \label{tab:effi}
  \begin{tabular}{@{}lcccccc@{}}
    \toprule
    Method & TrajCL & TrjSR & CLEAR & T-JEPA & BLUE & HiT-JEPA \\
    \midrule
    Time (s) & 0.196 & 0.476 & 0.292 & 0.303 & 0.166 & 0.341 \\
    \bottomrule
  \end{tabular}
  \vspace{-1. em}
\end{table}

\subsection{Visualizations of HiT-JEPA embeddings.}
HiT-JEPA encodes and predicts trajectory information only in the representation space, making it more difficult than generative models such as MAE~\cite{he2022masked} to evaluate the learned representation quality at the data level. To assess and gauge the validity of the representations of HiT-JEPA, we project the encoded $S{'}^{(1)}$ from $E_\theta^{(1)}$ (on full trajectories) and predicted $\widetilde{S}{'}^{(1)}$ from $D_{\phi}^{(1)}$ (on masked trajectories) back onto the hexagonal grid at their GPS coordinates for visual comparisons. \looseness -1

First, we freeze the context encoders and predictors across all levels in a pre-trained HiT-JEPA. Then we encode and predict the masked trajectory representations to simulate the training process, and encode the full trajectory representations to simulate the inference process. Next, we concatenate and tune a 2-layer MLP for each of the representations to decode to the hexagonal grid cell embeddings to which they belong. We denote the decoded predicted masked trajectory representations as $S_{1}$ and the decoded encoded full trajectory representations as $S_{2}$. Finally, for each trajectory position, we search for the $k$ most similar embeddings in the spatial region embedding set $\mathcal{H}$ and retrieve their hexagonal cell IDs. We choose $k=3$ in our visualizations.



\begin{figure}[htbp] 
  \centering
      \vspace{-.5 em}
  \begin{subfigure}[b]{0.9\linewidth} 
    \centering
    \includegraphics[height=5cm, width=.8\linewidth, keepaspectratio]{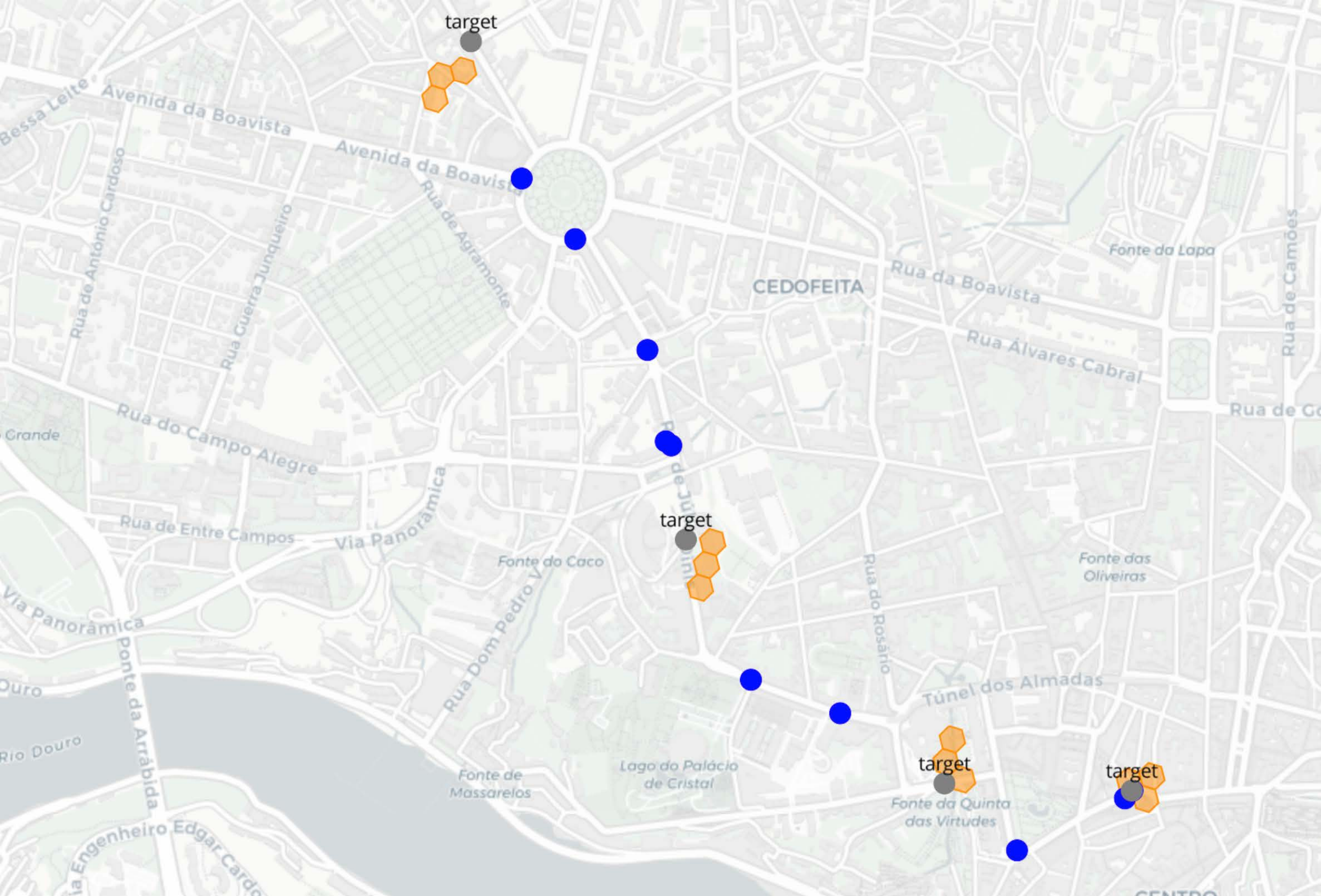}
    \caption{Predicted masked points}
    \vspace{-1 em}
    \label{fig:first}
  \end{subfigure}
  
  \vspace{1em} 
  
  \begin{subfigure}[b]{0.9\linewidth}
    \centering
    \includegraphics[height=5cm, width=.8\linewidth, keepaspectratio]{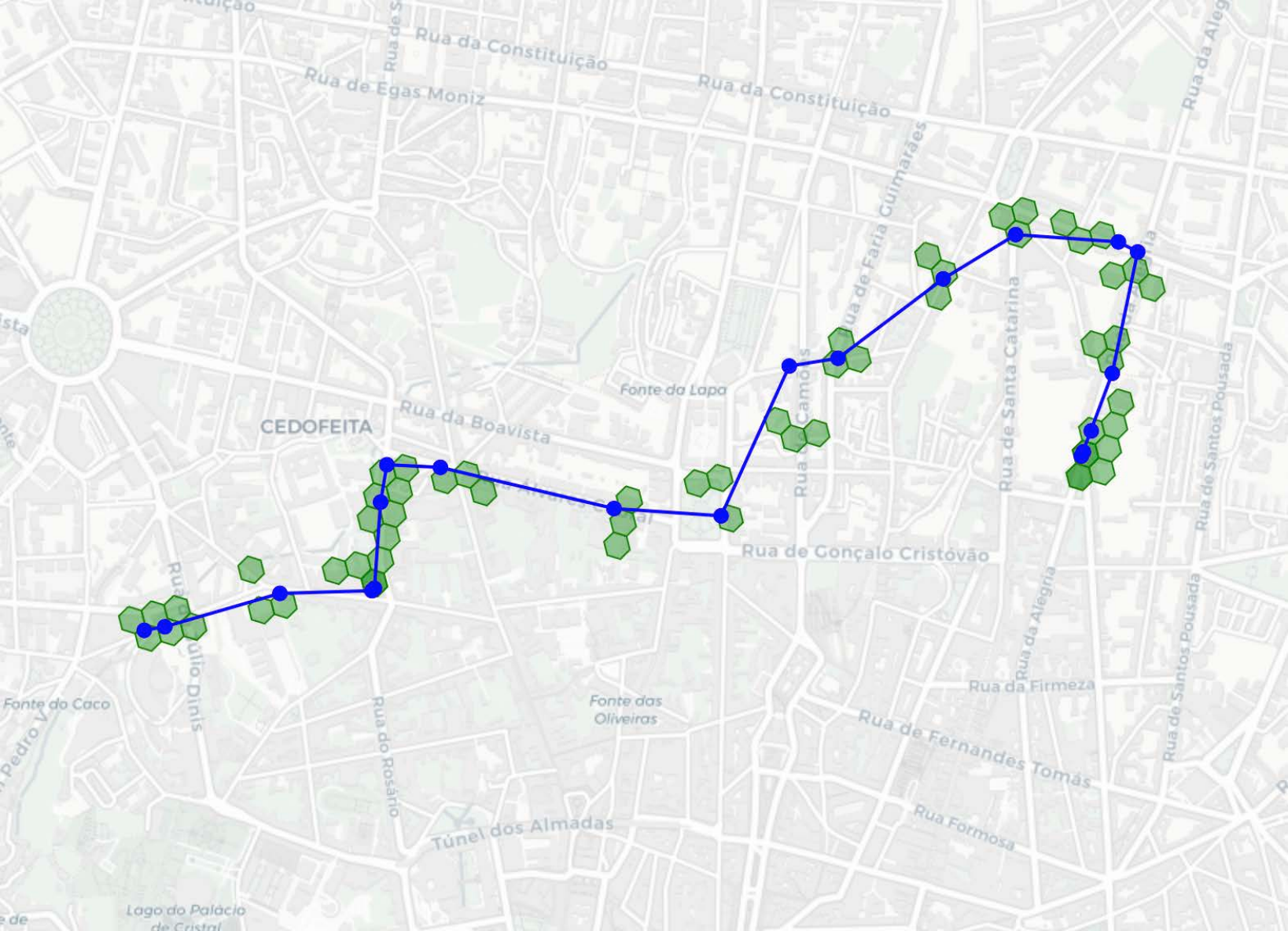}
    \caption{Encoded full trajectory}

    \vspace{-1 em}
    \label{fig:second}
  \end{subfigure}

  \caption{%
    Visualizations of decoded learned trajectory representations by HiT-JEPA on hexagonal cells: (a) \textcolor{blue}{blue} points are sampled trajectory points, \textcolor{gray}{gray} points are masked trajectory points labeled with "target", and \textcolor{orange}{orange} hexagons are projected predictions. 
    (b) \textcolor{blue}{blue} points are full trajectory points, \textcolor{green}{green} hexagons are projected encoded representations.
  }
  \vspace{-1.8em} 
  \label{fig:vis_main}
\end{figure}

\begin{figure*}[!t]
\centering
\vspace{-1 em}
\includegraphics[width=.85\linewidth]{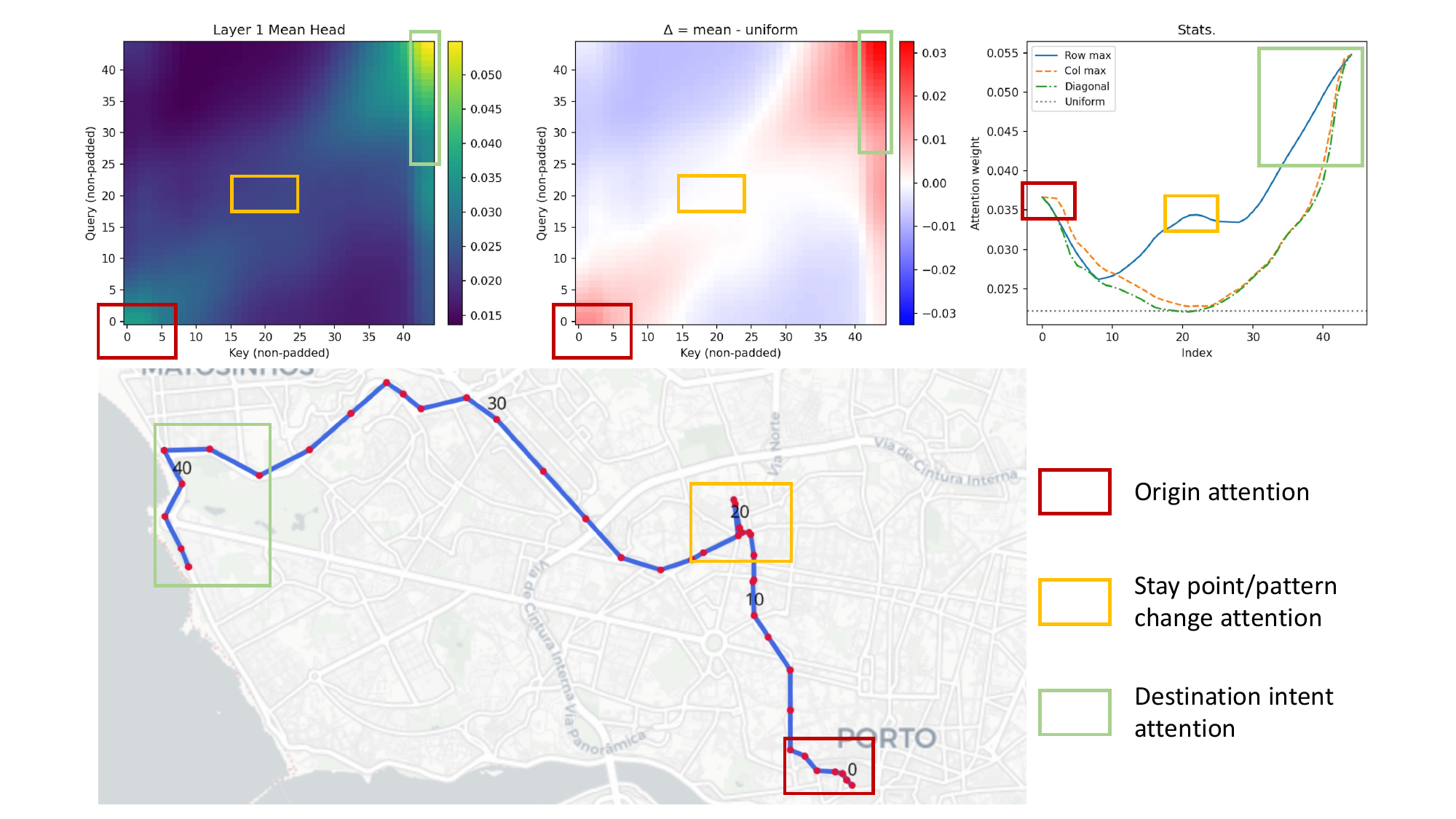}
\setlength{\abovecaptionskip}{-.5 em}
\caption{
A case study of hierarchical semantic information captured by HiT-JEPA. \textbf{(Top-Left)} The raw attention map visualizes the absolute attention weights, showing the overall intensity distribution. \textbf{(Top-Mid)} The deviation heatmap by displaying areas of active focus (red) versus suppression (blue) relative to the mean attention value. \textbf{(Top-Right)} The statistical profiles quantify the peak attention intensity at each time step. \textbf{Bottom} The corresponding physical trajectory with index labeled every 10 steps, where colored boxes spatially ground the salient attention regions identified in the top row.
}
\vspace{-1.5 em}
\label{fig:interp_main}
\end{figure*}

\begin{figure*}[!ht]
\centering
\includegraphics[width=.75\linewidth]{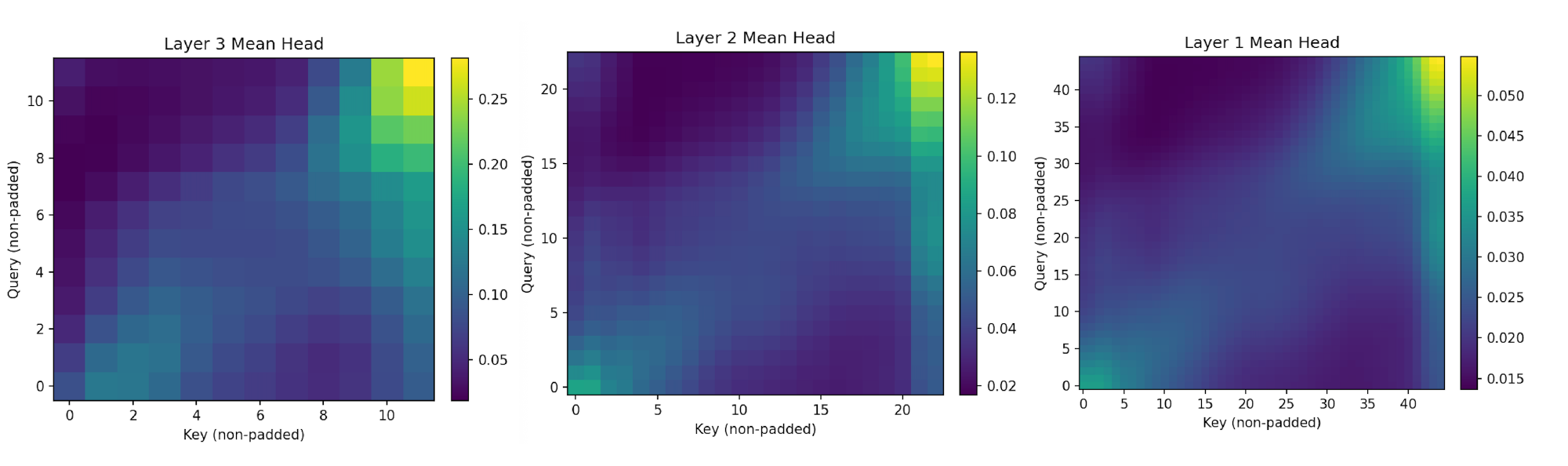}
\setlength{\abovecaptionskip}{-.5 em}
\caption{
Averaged attention weight visualizations at each JEPA layer. Left to right: $A^{(3)}$ to $A^{(1)}$.}

\vspace{-1.5 em}
\label{fig:3l}
\end{figure*}

\begin{table}[ht]
  \centering
  \footnotesize
  \setlength{\tabcolsep}{3pt}
  \renewcommand{\arraystretch}{0.8}
  \caption{Ablation Study of HiT-JEPA on Porto}
  \label{tab:ablation}
  \resizebox{0.48\textwidth}{!}{%
    \begin{tabular}{lccccc}
      \toprule
      \multicolumn{6}{c}{\textbf{Varying DB Size $\lvert D\rvert$}} \\
      \cmidrule(lr){1-6}
      \textbf{Model} & \textbf{20\%} & \textbf{40\%} & \textbf{60\%} & \textbf{80\%} & \textbf{100\%} \\ 
      \midrule
      HiT\_emb & 106.568 & 209.746 & 297.919 & 394.111 & 497.064 \\
      HiT\_sl & 1.031 & 1.061 & 1.066 & 1.077 & 1.091 \\
      HiT\_no\_at & 1.026 & 1.049 & 1.054 & 1.062 & 1.069 \\
      HiT\_rect & 1.032 & 1.062 & 1.069 & 1.080 & 1.093 \\
      \textbf{HiT-JEPA} & \textbf{1.026} & \textbf{1.043} & \textbf{1.048} & \textbf{1.058} & \textbf{1.065} \\
      \midrule
      \multicolumn{6}{c}{\textbf{Downsampling Rate $\rho_{s}$}} \\
      \cmidrule(lr){1-6}
      \textbf{Model} & \textbf{0.1} & \textbf{0.2} & \textbf{0.3} & \textbf{0.4} & \textbf{0.5} \\ 
      \midrule
      HiT\_emb & 569.322 & 706.831 & 1004.246 & 2047.699 & 2171.331 \\
      HiT\_sl & 1.378 & 2.659 & 5.626 & 14.123 & \textbf{26.875} \\
      HiT\_no\_at & 1.405 & 2.867 & 5.761 & 17.143 & 27.324 \\
      HiT\_rect & 1.508 & 3.054 & 7.735 & 18.912 & 36.768 \\
      \textbf{HiT-JEPA} & \textbf{1.369} & \textbf{2.624} & \textbf{5.541} & \textbf{13.773} & 28.806 \\
      \midrule
      \multicolumn{6}{c}{\textbf{Distortion Rate $\rho_{d}$}} \\
      \cmidrule(lr){1-6}
      \textbf{Model} & \textbf{0.1} & \textbf{0.2} & \textbf{0.3} & \textbf{0.4} & \textbf{0.5} \\ 
      \midrule
      HiT\_emb & 502.259 & 503.876 & 506.333 & 507.738 & 507.082 \\
      HiT\_sl & 1.088 & 1.099 & 1.120 & 1.100 & 1.137 \\
      HiT\_no\_at & 1.079 & 1.095 & 1.105 & 1.093 & 1.120 \\
      HiT\_rect & 1.095 & 1.111 & 1.123 & 1.122 & 1.124 \\
      \textbf{HiT-JEPA} & \textbf{1.074} & \textbf{1.077} & \textbf{1.085} & \textbf{1.093} & \textbf{1.119} \\
      \bottomrule
    \end{tabular}
  }
  \vspace{-1.em}
\end{table}

Fig. \ref{fig:first} shows the comparisons between decoded cells (\textcolor{orange}{orange} hexagons) and masked points (\textcolor{gray}{gray} points) labeled as ``targets''. The decoded locations lie in close proximity to their corresponding masked targets, confirming that the model effectively learns accurate representations for masked points during training.
Fig. \ref{fig:second} overlays the decoded cells \textcolor{green}{green} hexagons) on each \textcolor{blue}{blue} trajectory point, demonstrating that the model can encode each point with even greater accuracy with access to the full trajectory during inference. We further visualize the clustering of HiT-JEPA embeddings in \ref{cluster}. \looseness -1

\subsection{Hierarchical attention interpretation.}


From Fig. \ref{fig:interp_main}, by corroborating the raw attention map, deviation heatmap, and statistical profiles, we identify three distinct semantic phases localized within the bottom-left, top-right, and middle regions around the 20\textsuperscript{th} trajectory point. 
These phases correspond to the peak intensity of attention allocated to specific trajectory segments: the origin anchoring (\textcolor{red}{red} boxes), the local attention peak triggered by a pattern change (\textcolor{orange}{orange} boxes), and the destination intent (\textcolor{green}{green} boxes). This spatial-semantic alignment confirms that HiT-JEPA successfully learns the critical semantic waypoints from raw GPS tracks and verifies the interpretability of its learned representations.
By comparing the raw attention weights across 3 JEPA layers in Fig. \ref{fig:3l}, it is obvious to discern a coarse-to-fine attention evolution, where the $A^{3}$ layer highly summarizes the trajectory origin-destination patterns and is fused into lower layers with more smoothed local details. This validates that HiT-JEPA learns consistent trajectory semantics through the hierarchical interactions while preserving distinct layer-specific granularity. \looseness -1

\begin{figure*}[!ht]
\centering
\includegraphics[width=.95\linewidth]{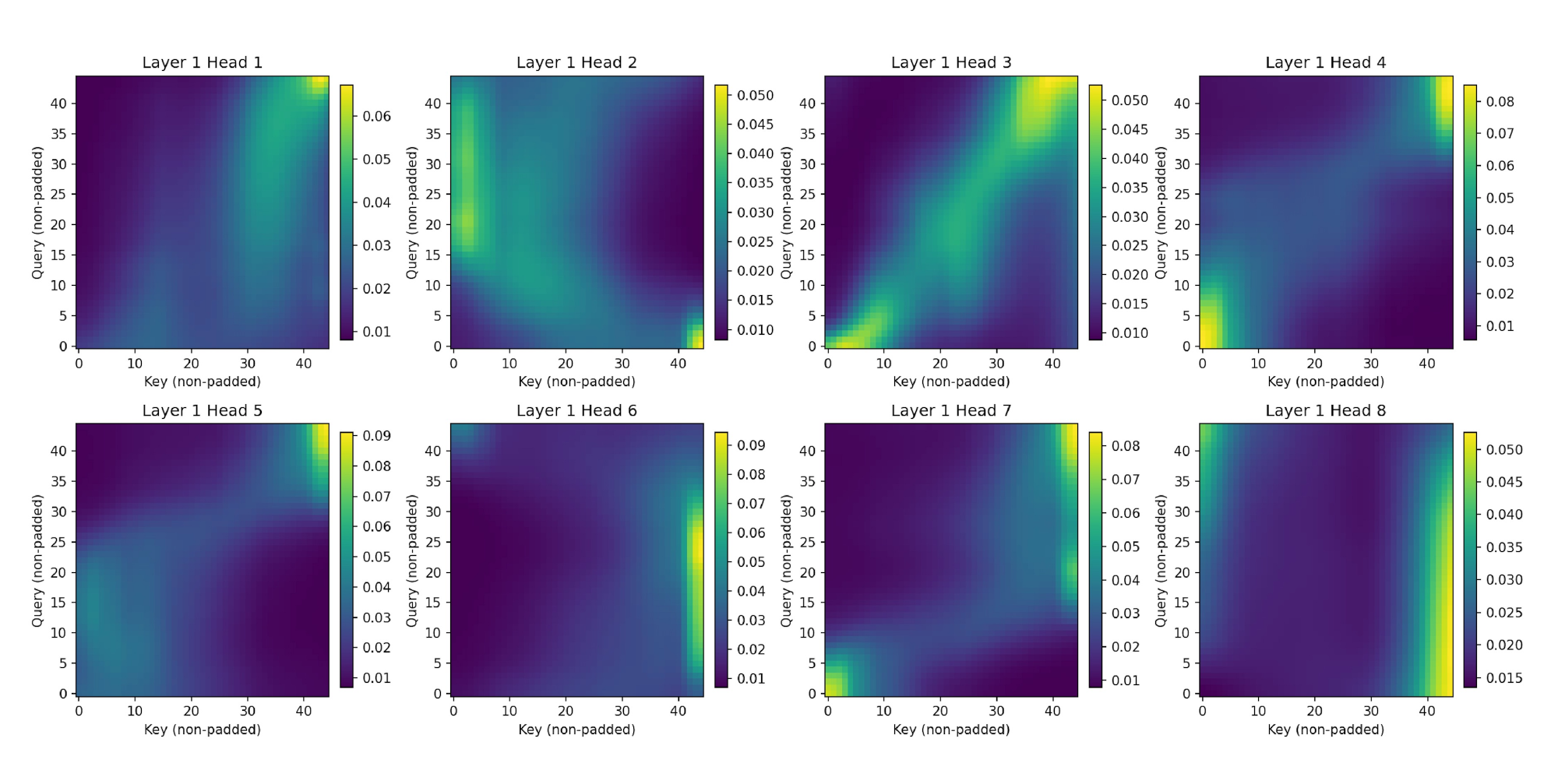}
\caption{
Visualization of each of the 8 attention heads at the JEPA layer $A^{(1)}$.
}
\vspace{-3 pt}
\label{fig:multihead}
\end{figure*}

The attention weights in Fig. \ref{fig:multihead} demonstrate functional specialization among attention heads. For example, Head 3 focuses on local kinematics, while Heads 2, 4, 6, and 8 act as global anchors that attend to long-term trajectory semantics. This diversity ensures a comprehensive representation that integrates fine-grained motion dynamics with high-level trip intent.

\subsection{Ablation Study}

We study the effect of removing the key designs in HiT-JEPA. We compare HiT-JEPA with 4 variants: 1) \textbf{HiT\_emb} which replaces the hierarchical interaction method from attention upsampling to directly concatenate the upsampled encoder embeddings between $S{'}^{(l)}$ and $S{'}^{(l-1)}$. 2) \textbf{HiT\_single\_layer} where we only level $l=1$ to train and predict. 3)  \textbf{HiT\_no\_attn} with no hierarchical interactions between each pair of successive layers. 4) \textbf{HiT\_rect} with the spatial location tokenization method changed to rectangular grid cells. We train these variants and conduct self-similarity experiments on Porto.  \looseness -1
Table \ref{tab:ablation} shows the comparisons between HiT-JEPA and its variants. The performance drops without any key designs, especially for HiT\_emb, as directly concatenating the embedding from the previous layers causes representation collapse. Results from the other two variants demonstrate that in our model design, even though each layer of $\mathrm{JEPA}^l$ can learn individually, the hierarchical interactions bind different levels into a cohesive multi-scale structure. We also conduct hyperparameter sensitivity analyses on key model settings in \ref{hype_analysis}.\looseness -1


\section{Conclusion}
HiT-JEPA introduces a unified three-layer hierarchy that captures point-level fine-grained details, intermediate trajectory patterns, and high-level trajectory semantics within a single self-supervised framework. By leveraging a Hierarchical JEPA, it enables more powerful trajectory feature extraction in the representation space and produces cohesive multi-granular embeddings. Extensive evaluations on diverse urban and maritime trajectory datasets show that HiT-JEPA outperforms single-scale self-supervised methods in trajectory similarity computation, particularly in zero-shot generalization and downstream fine-tuning, validating its effectiveness and robustness for real-world, large-scale trajectory modeling.

Currently, HiT-JEPA realizes hierarchical interaction through attention-weight upsampling and fusion, an approach tailored to Transformer-based JEPA models. A promising direction for future work is to generalize it into a unified framework in which architectures such as Mamba or LSTMs can plug in their own hierarchy modules under a consistent multi-level paradigm. Beyond standalone retrieval, HiT-JEPA can also serve as an efficient retrieval backend for LLM-based agents, providing similarity-preserving trajectory embeddings at a scale and cost that language models cannot match. \looseness -1

\newpage
\begin{acks}
We acknowledge the resources and services from the National Computational Infrastructure (NCI), which is supported by the Australian Government.
\end{acks}

\bibliographystyle{ACM-Reference-Format}
\bibliography{sample-base}

\appendix
\section{Appendix}\label{app}

\subsection{Experimental Settings}\label{exp}

\subsubsection{Self-similarity}
For each query trajectory $q\in Q$, we create two sub-trajectories $q_{a}=\{p_{1}, p_{3}, p_{5}, \ldots\}$ containing the odd-indexed points and $q_{b}=\{p_{2}, p_{4}, p_{6}, \ldots\}$ even-indexed points of $q$. We separate them by putting $q_{a}$ into the query set $Q$ and putting $q_{b}$ into the database $D$, with the rest of the trajectories in $D$ randomly filled from the testing set. Each $q_{a}$ and $q_{b}$ pair exhibits similar overall patterns in terms of shape, length, and sampling rate. We apply HiT-JEPA context-encoders to both query and database trajectories, compute pairwise similarities, and sort the results in descending order. Next, we report the mean rank of each $q_{b}$ when retrieved by its corresponding query $q_{a}$: ideally, the true match appears at rank one. We choose $\{20\%, 40\%, 60\%, 80\%, 100\%\}$ of the total database size $|D|$ for evaluation. To further evaluate the robustness of learned trajectory representations, we also apply down-sampling and distortion on $Q$ and $D$. Specifically, we randomly mask points (with start and end points kept) with down-sampling probability $\rho_{s}$ and shift the point coordinates with distortion probability $\rho_{d}$. Both $\rho_{s}$ and $\rho_{d}$ represent the number of points to be down-sampled or distorted, ranging from $\{0.1, 0.2, 0.3, 0.4, 0.5\}$.

For the convenience of comparing results under these settings together, we denote meta ratio $R_{i} = \{\mathrm{|}D\mathrm{|}_{i}, {\rho_{s}}_{i}, {\rho_{d}}_{i}\}$ and compare the \textbf{mean rank} of all models at each $R_{i}$ on each dataset, smaller values are better. \looseness -1

\subsection{Hyperparameter Analysis}\label{hype_analysis}
We analyze the impact of two sets of hyperparameters with the implementation and experimental settings.

\begin{figure}[htbp]
  \centering
  \begin{subfigure}[b]{0.32\linewidth}
    \centering
    \includegraphics[width=\linewidth]{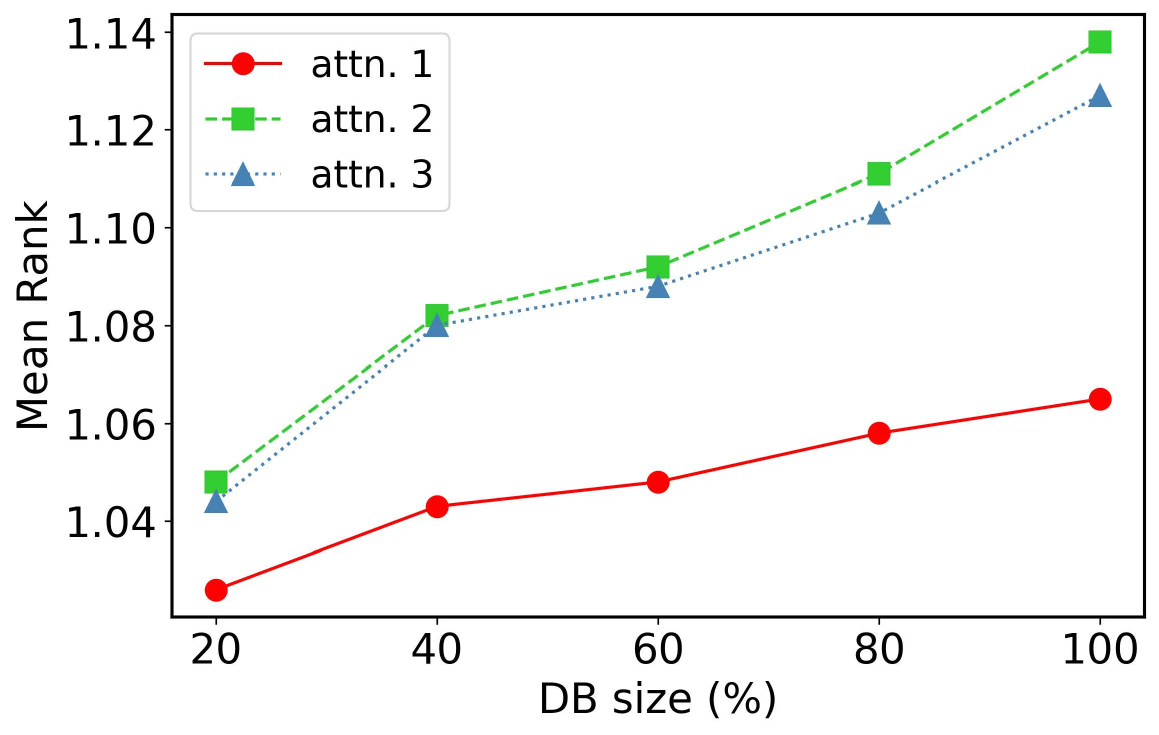}
    \caption{$\mathrm{|}D\mathrm{|}$ (20\%\(\sim\)100\%)}
    \label{fig:attn_variants}
  \end{subfigure}
  \hfill
  \begin{subfigure}[b]{0.32\linewidth}
    \centering
    \includegraphics[width=\linewidth]{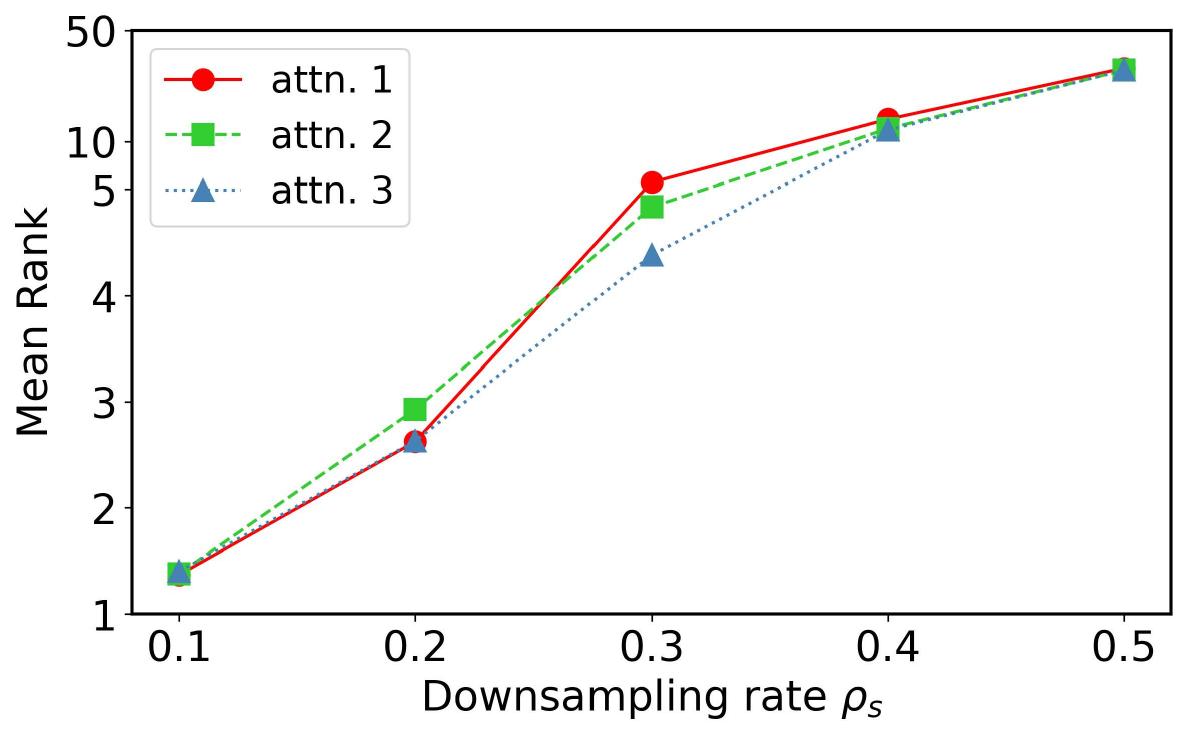}
    \caption{$\rho_{s}$ (0.1\(\sim\)0.5)}
    \label{fig:downsampling}
  \end{subfigure}
  \hfill
  \begin{subfigure}[b]{0.32\linewidth}
    \centering
    \includegraphics[width=\linewidth]{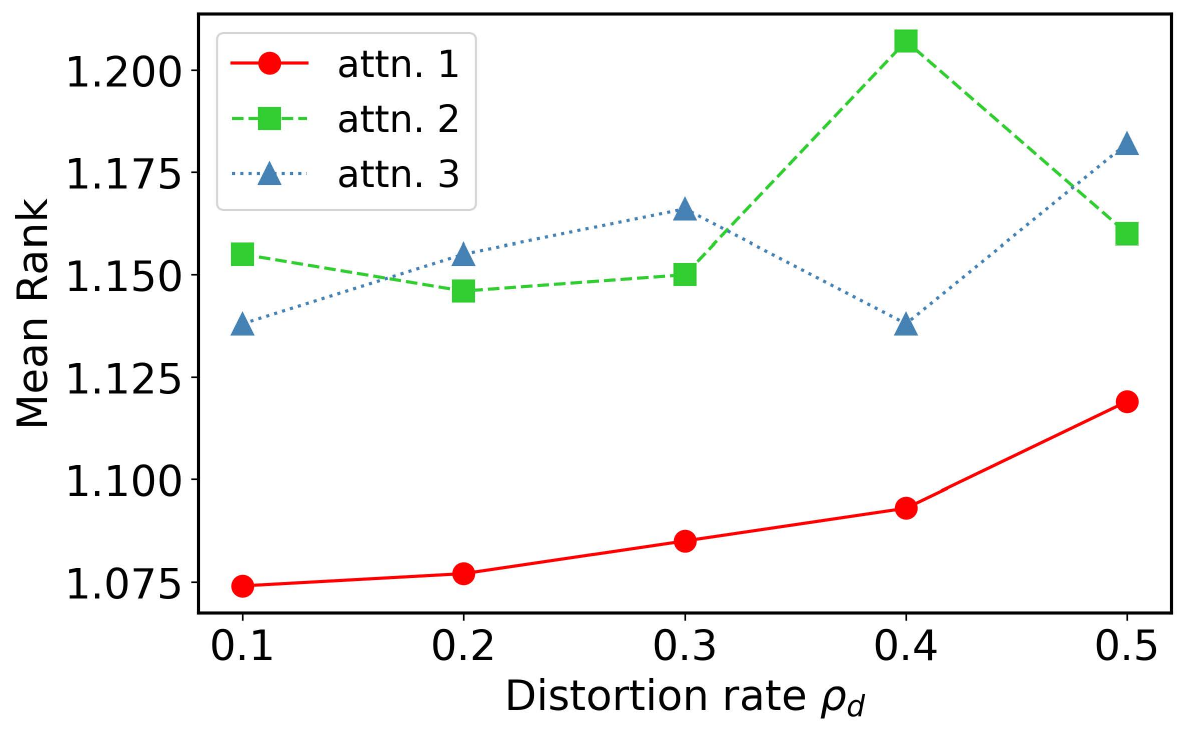}
    \caption{$\rho_{d}$ (0.1\(\sim\)0.5)}
    \label{fig:distortion}
  \end{subfigure}
  
  \caption{  
    Effect of different numbers of attention layers at each abstraction level.
  }
  \label{fig:attn_layers}
\end{figure}

\textbf{Number of attention layers at each abstraction level.} We change the number of Transformer encoder layers for each level to 2 and 3, then compare them with the default setting (1 layer) for self-similarity search with varying $\mathrm{|}D\mathrm{|}$, $\rho_{s}$ and $\rho_{d}$ on Porto. From Fig.~\ref{fig:attn_layers}, we can find that with only 1 attention layer, we can achieve the lowest mean ranks for all settings. This is due to higher chances of overfitting with more attention layers.

\begin{figure}[htbp]
  \centering
  \begin{subfigure}[b]{0.32\linewidth}
    \centering
    \includegraphics[width=\linewidth]{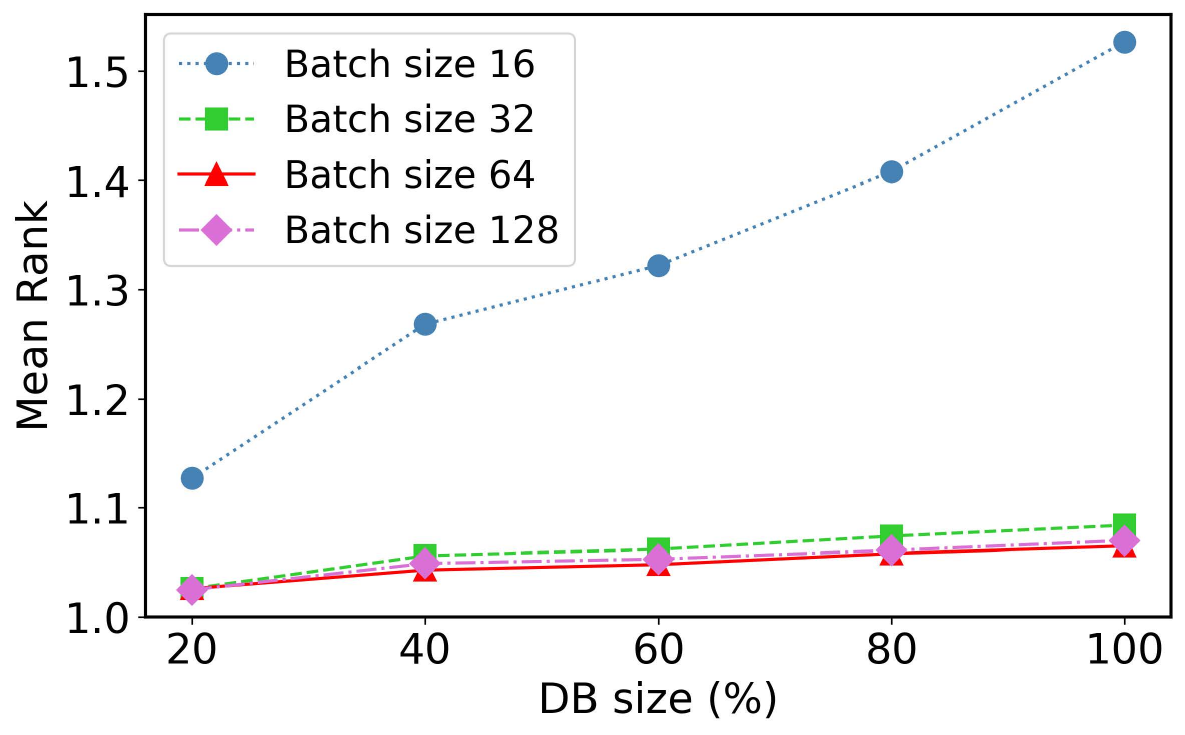}
    \caption{$\mathrm{|}D\mathrm{|}$ (20\%\(\sim\)100\%)}
    \label{fig:attn_db}
  \end{subfigure}
  \hfill
  \begin{subfigure}[b]{0.32\linewidth}
    \centering
    \includegraphics[width=\linewidth]{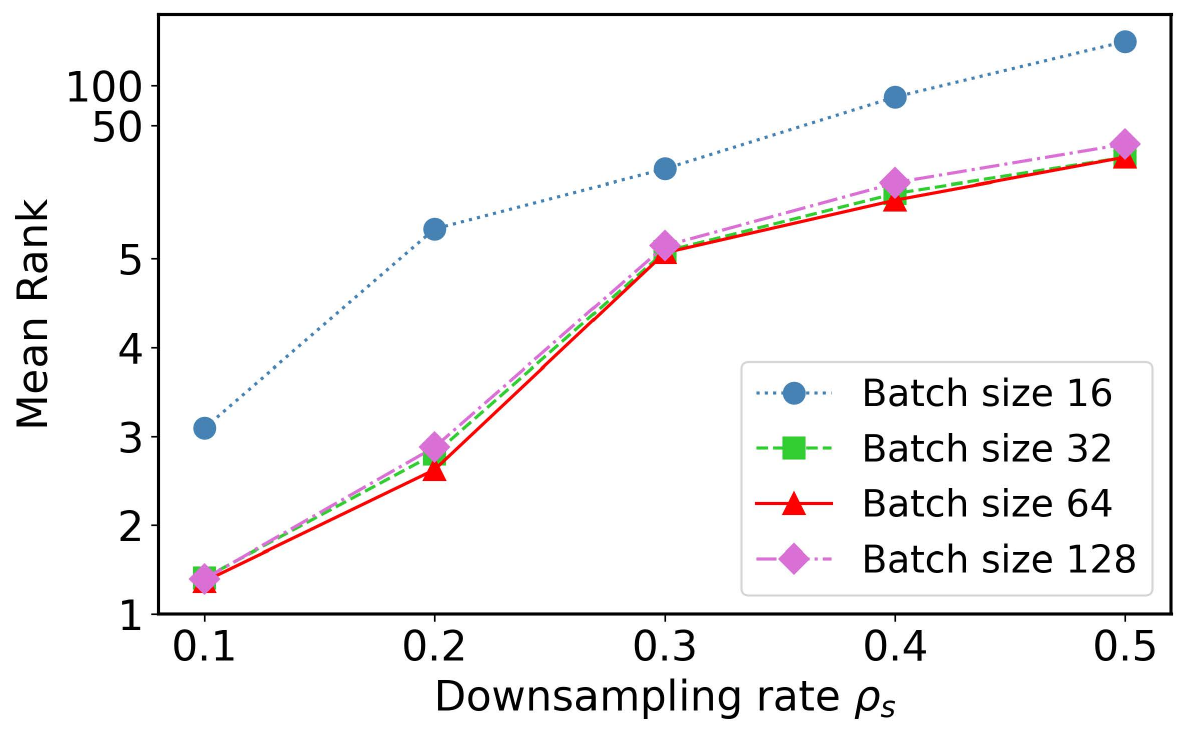}
    \caption{$\rho_{s}$ (0.1\(\sim\)0.5)}
    \label{fig:attn_downsampling}
  \end{subfigure}
  \hfill
  \begin{subfigure}[b]{0.32\linewidth}
    \centering
    \includegraphics[width=\linewidth]{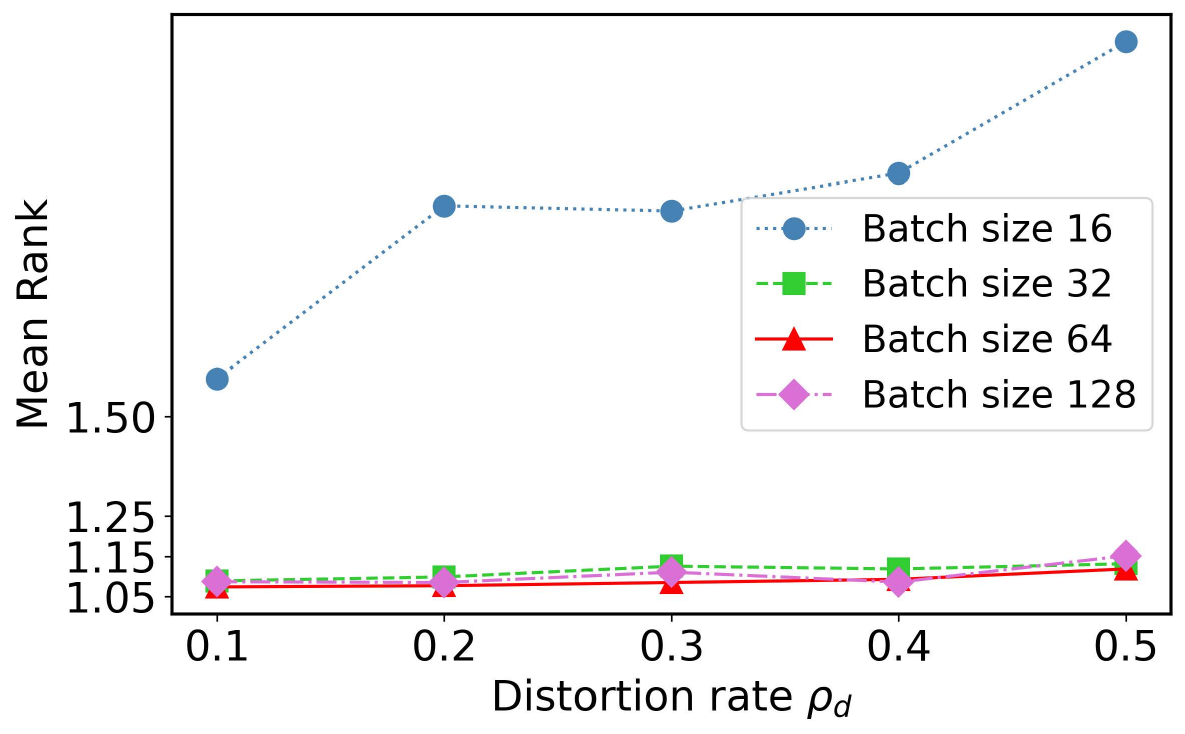}
    \caption{$\rho_{d}$ (0.1\(\sim\)0.5)}
    \label{fig:attn_distortion}
  \end{subfigure}
  
  \caption{  
    Effect of different batch sizes.
  }
  \label{fig:batch_size}
\end{figure}

\textbf{Weighting coefficient for the multi-level loss $\mathcal{L}$.} The values of the loss weighting coefficients $\lambda$, $\mu$, and $\nu $ are carefully tuned. In Table \ref{tab:coe-D}, \ref{tab:coe-DOW}, and \ref{tab:coe-DIS}, we compare our selected coefficients with other 3 sets of parameters in a wide range on the Porto dataset. From the tables, 

we can see that HiT-JEPA is robust against various loss combinations. Even though the loss coefficients with $\lambda$, $\mu$, and $\nu$ equal to 0.33, 0.33, and 0.33 perform better on the downsampling experiment, our selected combination with $\lambda=0.05$, $\mu=0.15$, and $\nu=0.8$ still learns overall the most accurate, stable, and consistent results across all experimental settings.

\begin{table}[htbp]
\small
  \centering
    \caption{Loss weighting coefficients for varying DB sizes $|D|$.}
  \begin{tabular}{cccrrrrr}
    \toprule
    $\lambda$ & $\mu$ & $\nu$ & 20\% & 40\% & 60\% & 80\% & 100\% \\
    \midrule
    0.1  & 0.2  & 0.7  & 1.026 & 1.050 & 1.056 & 1.067 & 1.079 \\
    0.33 & 0.33 & 0.33 & 1.036 & 1.072 & 1.080 & 1.102 & 1.120 \\
    0.6  & 0.3  & 0.1  & 1.035 & 1.063 & 1.066 & 1.079 & 1.099 \\
    \textbf{0.05} & \textbf{0.15} & \textbf{0.8} &
      \textbf{1.026} & \textbf{1.043} & \textbf{1.048} & \textbf{1.058} & \textbf{1.065} \\
    \bottomrule
  \end{tabular}

  \label{tab:coe-D}
\end{table}
\begin{table}[htbp]
\small
  \centering
    \caption{Loss weighting coefficients for varying downsampling rates $\rho_{s}$.}
  \begin{tabular}{cccrrrrr}
    \toprule
    $\lambda$ & $\mu$ & $\nu$ & 0.1 & 0.2 & 0.3 & 0.4 & 0.5 \\
    \midrule
    0.1  & 0.2  & 0.7  & \textbf{1.334} & 2.844 & 5.868 & 13.864 & 25.009 \\
    0.33 & 0.33 & 0.33 & 1.393 & 2.664 & \textbf{4.616} & \textbf{11.210} & \textbf{20.730} \\
    0.6  & 0.3  & 0.1  & 1.449 & 2.763 & 5.629 & 14.104 & 23.985 \\
    \textbf{0.05} & \textbf{0.15} & \textbf{0.8} &
      1.369 & \textbf{2.624} & 5.541 & 13.773 & 28.806 \\
    \bottomrule
  \end{tabular}

  \label{tab:coe-DOW}
\end{table}
\begin{table}[htbp]
\small
  \centering
    \caption{Loss weighting coefficients for varying distortion rates $\rho_{d}$.}
  \begin{tabular}{cccrrrrr}
    \toprule
    $\lambda$ & $\mu$ & $\nu$ & 0.1 & 0.2 & 0.3 & 0.4 & 0.5 \\
    \midrule
    0.1  & 0.2  & 0.7  & 1.081 & 1.109 & 1.095 & 1.115 & 1.135 \\
    0.33 & 0.33 & 0.33 & 1.130 & 1.138 & 1.133 & 1.152 & 1.196 \\
    0.6  & 0.3  & 0.1  & 1.092 & 1.107 & 1.133 & 1.117 & 1.191 \\
    \textbf{0.05} & \textbf{0.15} & \textbf{0.8} &
      \textbf{1.074} & \textbf{1.077} & \textbf{1.085} & \textbf{1.093} & \textbf{1.119} \\
    \bottomrule
  \end{tabular}

  \label{tab:coe-DIS}
\end{table}

\subsection{Representation visualization via clustering}\label{cluster}

\begin{figure}[!ht]
\centering
\includegraphics[width=.9\linewidth]{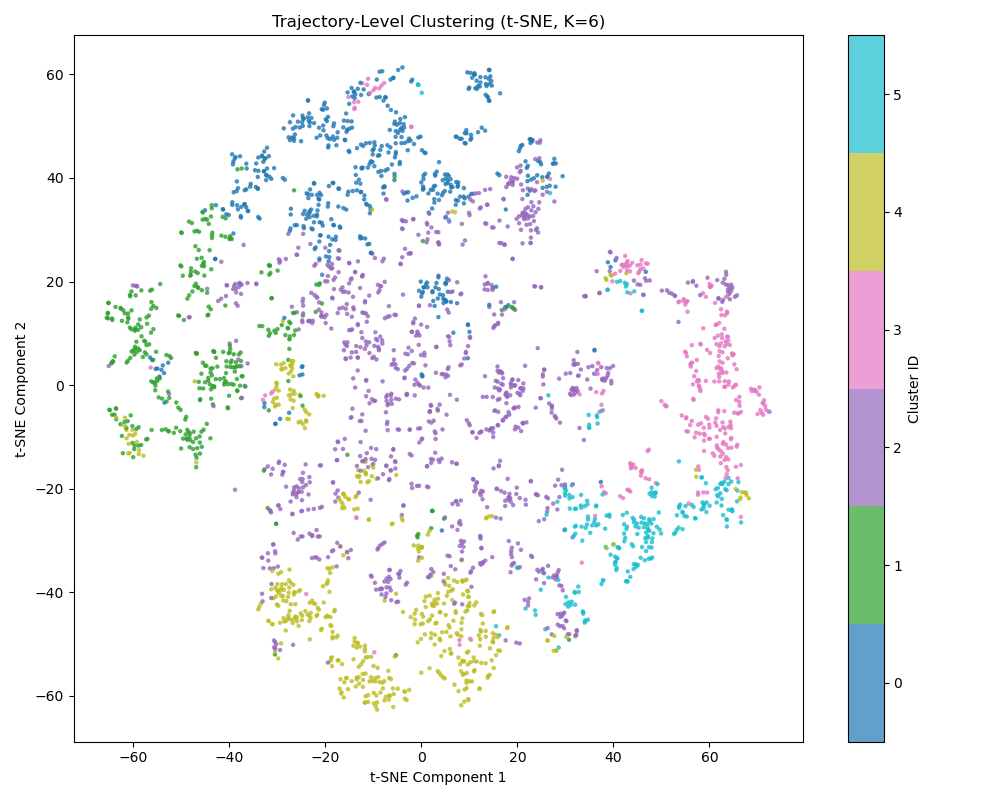}
\caption{
t-SNE visualization of trajectory embeddings.
}
\label{fig:clustering}
\end{figure}

We cluster and visualize the embedding of 3000 random trajectories in Porto in Fig. \ref{fig:clustering}. We use a K-Means Clusterer with a number of cluster centers $K=6$ acquired from the Elbow Method. We can find that, although the boundaries between clusters are soft without a specific self-clustering design in recent clustering methods~\cite{yao2017trajectory,fang20212}, distinct semantic groups are clearly visually discernible. This demonstrates the strong potential of HiT-JEPA, which is trained on regression loss, to be fine-tuned to generalize to multiple trajectory tasks. \looseness -1



\newpage





\subsection{Ethics Statement}
We claim that we adhere to the ACM Publications Policies. All the datasets used in the manuscript are publicly available with no user information revealed. HiT-JEPA encodes the trajectory location information in hexagonal cell tokens, where exact GPS traces are blurred. And such tokens are the only input to our model, thereby preventing any leakage of precise location data. The code for all baselines is publicly available and used under their respective licenses.

\end{document}